\documentclass[journal]{IEEEtran}
\IEEEoverridecommandlockouts
\usepackage{cite}
\usepackage{amsmath,amssymb,amsfonts}
\usepackage{algorithmic}
\usepackage{graphicx}
\usepackage{textcomp}
\usepackage{xcolor}
\usepackage{hyperref}
\usepackage{subcaption}
\usepackage{algorithm}
\hypersetup{hidelinks} 
\usepackage{caption}
\def\BibTeX{{\rm B\kern-.05em{\sc i\kern-.025em b}\kern-.08em
    T\kern-.1667em\lower.7ex\hbox{E}\kern-.125emX}}
\begin{document}

\title{Mobility-Aware Asynchronous Federated Learning with Dynamic Sparsification}

\author{\IEEEauthorblockN{Jintao Yan,~\IEEEmembership{Student Member,~IEEE,} Tan Chen,~\IEEEmembership{Graduate Student Member,~IEEE,} Yuxuan Sun,~\IEEEmembership{Member,~IEEE},\\ Zhaojun Nan,~\IEEEmembership{Member,~IEEE,} Sheng Zhou,~\IEEEmembership{Senior Member,~IEEE} and Zhisheng Niu,~\IEEEmembership{Fellow,~IEEE} \vspace{-2.5em}}\\
\thanks{J. Yan, T. Chen, Z. Nan, S. Zhou (Corresponding Author) and Z. Niu are with the Beijing National Research Center for Information Science and Technology, Department of Electronic Engineering, Tsinghua University, Beijing 100084, China. (email: \{yanjt22, chent21\}@mails.tsinghua.edu.cn, nzj660624@mail.tsinghua.edu.cn, \{sheng.zhou, niuzhs\}@tsinghua.edu.cn). 

Y. Sun is with the School of Electronic and Information Engineering, Beijing Jiaotong University, Beijing 100044, China. (e-mail: yxsun@bjtu.edu.cn).

An earlier version of this paper was presented in part at the IEEE International Conference on Computer Communications (INFOCOM) 2025 Workshop on Pervasive Network Intelligence for 6G Networks.}
}

\maketitle

\maketitle


\begin{abstract}
Asynchronous Federated Learning (AFL) enables distributed model training across multiple mobile devices, allowing each device to independently update its local model without waiting for others. However, device mobility introduces intermittent connectivity, which necessitates gradient sparsification and leads to model staleness, jointly affecting AFL convergence. This paper develops a theoretical model to characterize the interplay among sparsification, model staleness and mobility-induced contact patterns, and their joint impact on AFL convergence. Based on the analysis, we propose a mobility-aware dynamic sparsification (MADS) algorithm that optimizes the sparsification degree based on contact time and model staleness. Closed-form solutions are derived, showing that under low-speed conditions, MADS increases the sparsification degree to enhance convergence, while under high-speed conditions, it reduces the sparsification degree to guarantee reliable uploads within limited contact time. Experimental results validate the theoretical findings. Compared with the state-of-the-art benchmarks, the MADS algorithm increases the image classification accuracy on the CIFAR-10 dataset by $8.76\%$ and reduces the average displacement error in the Argoverse trajectory prediction dataset by $9.46\%$.

\end{abstract}


\section{Introduction}

The advent of 6G networks promises to support a wide range of new applications, including autonomous driving, smart cities, and the internet of things \cite{review1, review2}. These applications generate massive data and require efficient training of machine learning (ML) models \cite{review3}. Traditional centralized ML introduces privacy concerns and high latency. With increasingly powerful edge devices such as mobile phones, smart vehicles, and IoT sensors, it becomes feasible to shift the ML training process from centralized servers to these edge devices themselves. This shift paves the way for federated learning (FL), a promising framework that enables timely model training while preserving data privacy \cite{sun2020edge,FLreview}.

FL allows multiple edge devices to collaboratively train an ML model without sharing raw data \cite{DS6}. Each device performs local model training on its private dataset and uploads model updates to an edge server, which then aggregates them to refine the global model. However, practical FL deployment in wireless networks often requires dense edge servers to ensure sufficient coverage and timely model aggregation, leading to high infrastructure costs. To address this issue, many mobile devices in wireless networks, such as connected vehicles, unmanned aerial vehicles (UAVs), and portable 5G base stations, can themselves serve as edge servers \cite{MEET, V2Vedge3, UAV1}. These mobile edge servers (MESs) are often equipped with computational and storage capabilities to perform model aggregation and coordination. Leveraging such MESs, the FL framework can be flexibly deployed without relying solely on fixed infrastructure, thereby reducing deployment costs and enhancing scalability in dynamic or infrastructure-limited environments.

However, the mobility of both MES and edge devices leads to \emph{unstable and intermittent connections} \cite{MEET,mobD2D}. This challenges traditional synchronous FL, where the model aggregation process can start only after all scheduled devices have finished local training and model uploading. To overcome this limitation, asynchronous federated learning (AFL) has been proposed \cite{asy3, asy4, asy5}, enabling model aggregation to proceed without waiting for all scheduled devices.  AFL improves robustness against intermittent connectivity and enables continuous training despite frequent disconnections.

Although the asynchronous approach improves the flexibility of FL in mobile environments, it still faces significant challenges. The first challenge is \emph{model staleness}, which arises because devices perform local training based on outdated versions of the global model, caused by asynchronous aggregation and intermittent connections. This deviation from the current global model impairs convergence performance. Second, \emph{limited contact time and energy constraints} prevent full model uploads. To address this issue, sparsification techniques have been introduced, where only important gradients are transmitted to reduce communication overhead \cite{spar2, asy5}. However, this inevitably introduces sparsification errors, thereby degrading the quality of model updates. The mutual reinforcement of model staleness and sparsification errors exacerbates the degradation of AFL convergence.

While mobility causes intermittent connections, it also \emph{increases communication opportunities} between devices \cite{mobm1}, resulting in a twofold impact on AFL convergence \cite{MEET}. Existing studies \cite{asy4, asy5, mobasy} lack a rigorous analysis of how these factors jointly affect convergence. Furthermore, controlling the sparsification degree and energy consumption under highly dynamic environments is often neglected.
This work develops a theoretical model to characterize the impact of sparsification, model staleness and mobility-induced contact patterns on AFL convergence, and proposes a mobility-aware dynamic sparsification (MADS) algorithm based on the theoretical analysis. The main contributions are summarized as follows.

\begin{itemize}
\item Through convergence analysis, we find that reducing contact time leads to performance degradation due to increased sparsification error, while reducing inter-contact time improves performance by reducing model staleness. Furthermore, the analysis quantifies the dual impact of mobility on convergence, as increasing device speeds reduce both contact time and inter-contact time. At low speeds, the influence of reducing inter-contact time is dominant, so mobility improves convergence. At high speeds, the impact of the reduced contact time becomes more significant, so mobility degrades convergence.

\item Based on the theoretical analysis, we propose the MADS algorithm, which dynamically adjusts the sparsification degree according to the contact time and model staleness of each device. We derive closed-form solutions for optimal sparsification and power allocation, showing that under low-speed conditions, MADS increases the sparsification degree to transmit more gradient elements and accelerate convergence. Under high-speed conditions, MADS reduces the sparsification degree to ensure reliable model uploads within the limited contact time.

\item We evaluate the convergence performance of AFL on the CIFAR-10 and Argoverse trajectory prediction datasets. Simulations under varying contact times, inter-contact times, and device speeds validate the theoretical analysis. In addition, the proposed MADS algorithm improves the image classification accuracy on CIFAR-10 by $8.76\%$ and reduces the average displacement error (ADE) on Argoverse by $9.46\%$ compared to state-of-the-art benchmarks.

\end{itemize}

The rest of this paper is organized as follows. Section II reviews related work. Section III presents the system model. Section IV provides the convergence analysis, and Section V introduces the MADS algorithm. Experimental results are shown in Section VI, followed by conclusions in Section VII.

\section{Related Work}
\subsection{FL over Mobile Wireless Networks}

FL has been widely studied in wireless networks \cite{FLreview}, addressing challenges such as device scheduling and resource allocation \cite{DS3, DS4, DS9, DS6, DSsun}, quantization and sparsification \cite{CS4, CS5, CS1, CS2}, and aggregation algorithm design \cite{FLT2, chuanhuang}. However, most of these studies overlook the mobility of edge devices.

Recent studies have begun to investigate the impact of mobility on FL, focusing on performance optimization and convergence analysis in mobile networks \cite{mob1, VFL6, yjt, VFL7, mobfl, mobconv, chentan}. In \cite{mob1}, the time-varying channels and the dynamic network topologies resulting from mobility are considered. Ref. \cite{VFL6} points out that mobility degrades training data quality due to noise and distortion, and proposes a scheduling algorithm that selects high-quality data sources. Ref. \cite{yjt} leverages the vehicle-to-vehicle communications to enhance FL performance. Ref. \cite{VFL7} proposes a mobility- and channel-aware FL scheme to improve uploading success by leveraging mobility patterns.

Most of these works rely on system-level optimizations to mitigate mobility-induced degradation, with limited theoretical analysis of convergence performance. More recent studies \cite{mobfl, mobconv, chentan} characterize the impact of mobility based on convergence analysis. Ref. \cite{mobfl} introduces a hierarchical FL framework and shows that mobility degrades convergence due to increased model divergence. A user clustering strategy is proposed to stabilize updates. In contrast, Refs. \cite{mobconv, chentan} highlight positive effects of mobility. Ref. \cite{mobconv} shows that low mobility enhances FL by increasing data diversity, leading to better generalization. Ref. \cite{chentan} further decomposes the impact into data fusion, which accelerates convergence, and model shuffling, which may slow it down by introducing instability during aggregation.

\subsection{Mobile Edge Server}
MESs play a crucial role in the evolution of edge intelligence, offering computational, storage, and communication resources at the network edge, particularly in mobile environments \cite{moveintel}. They are deployed in various forms, such as connected vehicles, UAVs, portable 5G base stations, or mobile relay stations. Integrating MESs into the network architecture extends edge intelligence to dynamic environments, enhancing the scalability and efficiency of network services \cite{MEET}.

Recent studies have explored the benefits and challenges of MES in various domains. Refs. \cite{V2Vedge1, V2Vedge3, V2Vedge2} explore the role of vehicles as MESs in vehicular edge computing (VEC) systems. Ref. \cite{V2Vedge1} proposes a dynamic task offloading scheme where service vehicles act as edge servers, providing computation resources to nearby vehicles. Ref. \cite{V2Vedge3} extends this scheme by introducing multi-hop relays, ensuring that tasks can still be offloaded via intermediate vehicles when direct links are disrupted by mobility. Ref. \cite{V2Vedge2} combines fixed RSUs and mobile service vehicles to optimize offloading and resource allocation, with tasks dispatched via V2I or V2V, improving system efficiency and ensuring quality of service. In \cite{UAV1,UAV2}, UAVs are deployed as MESs, serving as both communication relays and computational resources. Their flexibility allows rapid deployment and adaptation to network changes, particularly in areas where fixed infrastructures are not available. \cite{UAVFL1, UAVFL2} explore UAVs as MESs in FL scenarios, acting as parameter servers to aggregate local models from devices. By establishing short-range line-of-sight links, UAVs reduce communication delays and mitigate straggler effects, thereby accelerating FL convergence.

\subsection{Asynchronous FL}
AFL has gained attention in recent years as a promising solution to the challenges posed by device heterogeneity and communication constraints in FL. Traditional synchronous FL approaches often suffer from the straggler effect, where the training process is delayed by slower devices. AFL addresses this by allowing devices to upload model updates independently, improving robustness under varying device availability and network conditions \cite{asy3}.

Early works on asynchronous stochastic gradient descent (SGD) \cite{asy1} show that it significantly reduces communication rounds compared to mini-batch SGD. Subsequent studies \cite{asy2, spar2} incorporate quantization and sparsification into asynchronous SGD. The above papers lay the foundation for AFL. Ref. \cite{asy3} formally introduces the AFL framework. This work rigorously analyzes the convergence rates for both strongly convex and nonconvex functions, demonstrating that their AFL achieves a convergence rate comparable to those of synchronous methods. Building on this, Ref. \cite{asy4} implements AFL in wireless networks and proposes three adaptive transmission scheduling algorithms to enhance learning efficiency. Ref. \cite{asy5} introduces a data-importance-aware scheduling policy that accounts for both data distribution and freshness. Ref. \cite{mobasy} leverages the relay opportunities brought about by the mobility of devices to improve the convergence rate of AFL. However, most existing studies lack a rigorous theoretical analysis of the impact of how contact time, inter-contact time, and device speed affect AFL convergence performance. Moreover, current methods overlook the need to adapt sparsification degrees in highly dynamic wireless environments.

\begin{figure}[t!]
\centering
\includegraphics[width=0.48\textwidth]{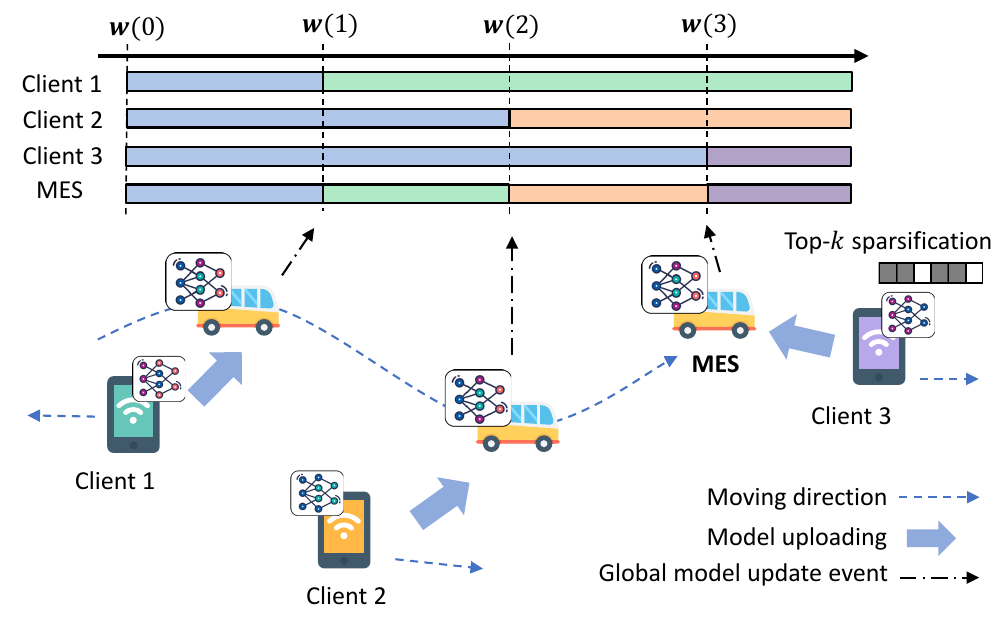}
\caption{The mobility-aware AFL framework.}
\label{system}
\end{figure}

\section{System Model}
\label{II}
\subsection{System Overview}

We consider an AFL framework over a mobile network, as illustrated in Fig. \ref{system}, where there is one MES and $N$ mobile devices, denoted as $\mathcal{N}$. The MES and the mobile devices collaboratively train an ML model $\boldsymbol{w}$.

Each mobile device $n\in\mathcal{N}$ holds a local dataset $\mathcal{D}_n$. For each data sample $\boldsymbol{d} \in \mathcal{D}_n$, a loss function evaluates the fitting performance of the local model parameter $\boldsymbol{w}_n$. The local loss function of device $n$ is defined as the average loss function over its dataset, i.e., $F_n(\boldsymbol{w}_n)\triangleq \frac{1}{|\mathcal{D}_n|} \sum_{\boldsymbol{d} \in \mathcal{D}_n}f(\boldsymbol{w}_n,\boldsymbol{d}).$ Considering all devices over the system, the global loss function of the FL task is defined as \begin{equation}F(\boldsymbol{w})=\frac{1}{N} \sum_{n \in \mathcal{N}}F_n(\boldsymbol{w}_n).\notag\end{equation} 
The objective of FL training is to minimize the global loss function by optimizing the global parameter $\boldsymbol{w}^*$ over $R$ training rounds, where $\boldsymbol{w}^*=\arg \min F(\boldsymbol{w})$, $\mathcal{R} = \{1,2,..., R\}$ denotes the set of training round indices, and the duration of each training round is denoted by $\delta$.

\subsection{Mobility Model}
We adopt the widely used inter-contact model to characterize device mobility patterns \cite{mobm1,mobm2,mobm3,mobm4}. The \emph{contact time} is defined as the duration during which a device is able to upload its gradients to the MES, and the \emph{inter-contact time} is defined as the interval between two consecutive contacts. In this model, the contact time $\tau_n$ between device $n$ and the MES follows an exponential distribution, i.e., $\tau_n\sim \exp(\frac{1}{c_n})$, where $c_n$ denotes the average contact time. The inter-contact time $t_n$ also follows an exponential distribution, i.e., $t_n\sim \exp(\frac{1}{\lambda_n})$,
where $\lambda_n$ denotes the average inter-contact time.

\subsection{The AFL Training Process}
The AFL training process is outlined in Algorithm \ref{Algo1}. At the start of the training, the MES initializes the global model $\boldsymbol{w}^{(0)}$. Each device initializes its local model $\boldsymbol{w}^{(0)}_n$, the cumulative gradients $\boldsymbol{g}^{(0)}_n$ and the local memory $\boldsymbol{e}^{(0)}_n$, where the local memory tracks the accumulated error introduced by sparsification \cite{spar1} \cite{spar2}. The FL training process then starts.
\subsubsection{On the Mobile Devices}
At each round $r\in\mathcal{R}$, every device $n\in\mathcal{N}$ computes the stochastic gradient of the local model, and updates the cumulative gradients as
\begin{equation}\boldsymbol{g}^{(r)}_n = \boldsymbol{g}^{(r-1)}_n + \eta \nabla f_n\left(\boldsymbol{w}^{(r-1)}_n,\mathcal{B}^{(r-1)}_n\right),\notag\end{equation}
where $\eta$ is the learning rate, $\mathcal{B}^{(r-1)}_n$ is a mini-batch randomly sampled from $\mathcal{D}_n$, and $f_n\left(\boldsymbol{w}^{(r-1)}_n,\mathcal{B}^{(r-1)}_n\right) = \frac{1}{|\mathcal{B}^{(r-1)}_n|}\sum_{\boldsymbol{d}\in\mathcal{B}^{(r-1)}_n} f\left(\boldsymbol{w}^{(r-1)}_n,\boldsymbol{d}\right)$ is the stochastic gradient. 

Let $\zeta_{n}^{(r)}$ denote whether the device contacts the MES. $\zeta_{n}^{(r)} = 1$ if device $n$ contacts the MES in round $r$. Otherwise, $\zeta_{n}^{(r)} = 0$. If device $n$ does not contact the MES in round $r$, it updates its local model using SGD as 
\begin{equation}\boldsymbol{w}^{(r)}_n = \boldsymbol{w}^{(r-1)}_n - \eta \nabla f_n\left(\boldsymbol{w}^{(r-1)}_n,\mathcal{B}^{(r-1)}_n\right).\notag\end{equation} 

If device $n$ contacts the MES in round $r$, it adds the cumulative gradients and the local memory to obtain the gradients to upload, denoted as
\begin{equation}\boldsymbol{x}^{(r)}_n = \boldsymbol{e}^{(r-1)}_n+\boldsymbol{g}^{(r)}_n.\notag\end{equation} 

Then, it sends the sparsified gradients $S_n^{(r)}(\boldsymbol{x}^{(r)}_n)$ to the MES for model aggregation, where $S_n^{(r)}(\cdot)$ denotes the sparsification operator introduced in Section \ref{quan}. The local memory is updated as $\boldsymbol{e}^{(r)}_n = \boldsymbol{x}^{(r)}_n - S_n^{(r)}(\boldsymbol{x}^{(r)}_n)$. After aggregation, it receives the aggregated global model from the MES, updates its local model, and resets the cumulative gradients.

\subsubsection{On the MES}
At each round $r\in \mathcal{R}$, the MES receives the sparsified gradients from the devices it contacts, and updates the global model as
\begin{equation} \boldsymbol{w}^{(r)} = \boldsymbol{w}^{(r-1)} - \frac1 N \sum_{n\in\mathcal{N}} \zeta_{n}^{(r)} S_n^{(r)}(\boldsymbol{x}^{(r)}_n).\notag\end{equation} 
Then, the MES broadcasts the updated global models to the devices it contacts.

Let $\kappa_n^{(r)}$ denote the most recent round before or at $r$ when device $n$ received the global model, with $\kappa_n^{(0)} = 0$. We define the \emph{model staleness} of device $n$ as 
$$\theta^{(r)}_n = r - \kappa^{(r)}_n,$$
which measures the number of rounds elapsed since the last reception of an updated global model from the MES.

\subsection{Sparsification Model}
\label{quan}
Due to the limited contact time between devices and the MES, we adopt the top-$k$ sparsification techniques \cite{spar1, spar2} to reduce the number of gradients to upload. This process ranks the gradients by their absolute values and then transmits only the top-$k$ gradients with the largest magnitudes, where $k$ denotes the sparsification degree. The remaining gradients are omitted from transmission.

After top-$k$ sparsification, the total number of bits required for upload equals the sum of the bits for representing the selected indices and the corresponding gradient values. Let $s$ denote the number of model parameters, and $k_n^{(r)}$ be the number of selected gradients for device $n$ in round $r$. Then, the total number of bits uploaded is given by $uk_n^{(r)} + k_n^{(r)}\log_2 s$, where $u$ denotes the number of bits used to represent each gradient value (e.g., $u = 32$ for float32). This number must not exceed the transmission capacity during the contact time, i.e., $uk_{n}^{(r)}+k_{n}^{(r)}\log_2 s \leq \tau^{(r)}_n A_{n}^{(r)}$,
where $A_n^{(r)}$ is the transmission rate and $\tau_n^{(r)}$ is the contact time of device $n$ in round $r$.

\begin{algorithm}[!t]	
    \caption{The AFL training process} 
    \label{Algo1}
	\begin{algorithmic}
        \STATE{Initialize $\boldsymbol{w}^{(0)}$, $\boldsymbol{w}^{(0)}_n$, $\boldsymbol{g}^{(0)}_n$, and $\boldsymbol{e}^{(0)}_n$, $\forall n \in \mathcal{N}$.}
	\FOR{$r = 1$ \textbf{to} $R$}
         \STATE{\textbf{On each mobile device $n$ \textbf{in} $\mathcal{N}$}:} 
         \begin{ALC@g}
         \STATE{$\boldsymbol{g}^{(r)}_n = \boldsymbol{g}^{(r-1)}_n + \eta \nabla f_n\left(\boldsymbol{w}^{(r-1)}_n,\mathcal{B}^{(r-1)}_n\right)$;}   
        \IF{$\zeta^{(r)}_n = 1$}  \STATE{$\boldsymbol{x}^{(r)}_n = \boldsymbol{e}^{(r-1)}_n+\boldsymbol{g}^{(r)}_n$}  
        \STATE{Send $S_n^{(r)}(\boldsymbol{x}^{(r)}_n)$ to the MES;} 
        \STATE{Receive $\boldsymbol{w}^{(r)}$ from the MES;}
        \STATE{$\boldsymbol{e}^{(r)}_n = \boldsymbol{x}^{(r)}_n - S_n^{(r)}(\boldsymbol{x}^{(r)}_n)$, $\boldsymbol{w}^{(r)}_n = \boldsymbol{w}^{(r)}$ and $\boldsymbol{g}^{(r)}_n = 0$;}
        \ELSE
        \STATE{$\boldsymbol{e}^{(r)}_n = \boldsymbol{e}^{(r-1)}_n$};
        \STATE{$\boldsymbol{w}^{(r)}_n = \boldsymbol{w}^{(r-1)}_n - \eta \nabla f_n\left(\boldsymbol{w}^{(r-1)}_n,\mathcal{B}^{(r-1)}_n\right)$.}
        \ENDIF
         \end{ALC@g}
        \STATE{\textbf{On the MES:}}
        \begin{ALC@g}
        \IF{$\sum_{n\in\mathcal{N}} \zeta_{n}^{(r)}=0$} 
        \STATE{Set $\boldsymbol{w}^{(r)} = \boldsymbol{w}^{(r-1)}$;}
        \ELSE
        \STATE{Receive $S_n^{(r)}(\boldsymbol{x}^{(r)}_n)$ from $\{n | \zeta_{n}^{(r)} = 1\}$;}
        \STATE{ $\boldsymbol{w}^{(r)} = \boldsymbol{w}^{(r-1)} -\frac1 N \sum_{n\in\mathcal{N}} \zeta_{n}^{(r)} S_n^{(r)}(\boldsymbol{x}^{(r)}_n)$;}
        \STATE{Send $\boldsymbol{w}^{(r)}$ to the devices it contacts.}
        \ENDIF
        \end{ALC@g}
	\ENDFOR
	\end{algorithmic}
\end{algorithm}
\section{Convergence Analysis}
This section presents the convergence analysis of the proposed AFL. Following the state-of-the-art literature \cite{asy5, spar2, DSsun}, we make the following assumptions:

\noindent \emph{Assumption 1:}
The local loss function $F_n(\boldsymbol{w})$ is $L$-smooth for each device $n \in \mathcal{N}$, i.e., 
\begin{equation}
\begin{aligned}&F_n(\boldsymbol{w}') - F_n(\boldsymbol{w})\leq \left < { \nabla F_n(\boldsymbol{w}), \boldsymbol{w}'-\boldsymbol{w} }\right >  + \frac {L}{2} \left \Vert{ \boldsymbol{w}'-\boldsymbol{w}}\right \Vert ^{2}.\notag
\end{aligned}
\end{equation}

\noindent \emph{Assumption 2:}
For every $n\in\mathcal{N}$ and $r\in\mathcal{R}$, the squared norm of the stochastic gradient is bounded by a constant $G^2$, i.e.,
\begin{equation}
\mathbb{E}\left\|\nabla f_n\left(\boldsymbol{w}^{(r)}_n,\mathcal{B}^{(r)}_n\right)\right\|^2\leq G^2.\notag
\end{equation}
This also implies a bound on the variance:
\begin{equation}
\mathbb{E}\left\|\nabla f_n\left(\boldsymbol{w}^{(r)}_n,\mathcal{B}^{(r)}_n\right) - \nabla F_n(\boldsymbol{w}^{(r)}_n)\right\|^2\leq \sigma^2,\notag
\end{equation}
where $\sigma$ is also a constant and $\sigma^2 \leq G^2$, and the expectation is taken over the randomness of SGD.

To facilitate convergence analysis, we adopt a virtual model-based framework. Unlike synchronous FL, where all devices update from the same global model and synchronize after each round, AFL allows devices to update independently and asynchronously, resulting in inconsistent model states across devices. This asynchrony complicates theoretical analysis due to the divergence between local and reference global models.

To address this, we introduce a virtual model that aggregates all historical local updates as if they were performed synchronously from a common starting point. By comparing the virtual model with the asynchronously updated local models, we derive an upper bound on their divergence in Lemma 1. This divergence bound is then used to establish the convergence bound of AFL in Theorem 1. Firstly, we define the virtual model as

\begin{equation}
\boldsymbol{v}^{(r)} = \boldsymbol{w}^{(0)} - \frac1 N \sum_{n\in\mathcal{N}}\sum_{j=0}^{r-1}\nabla f_n\left(\boldsymbol{w}_n^{(j)},\mathcal{B}_n^{(j)}\right).\notag
\end{equation}
Then, we derive the following lemma to bound the difference between the virtual and real models.

\noindent\textbf{Lemma 1.} \emph{Under the given assumptions, the difference between the virtual and real models are bounded by}
\begin{equation}
\begin{aligned}
&\frac{1}N\sum_{n\in \mathcal{N}}\mathbb{E}\left\|\boldsymbol{v}^{(r)}-\boldsymbol{w}_{n}^{(r)}\right\|^{2}\leq\frac{1}N\sum_{n\in \mathcal{N}} 2\left(\theta_n^{(r)}\right)^2\eta^2G^2 \\
&\frac{1}N\sum_{n\in \mathcal{N}}\left(5-\frac{3k_n^{(\kappa^{(r)}_n)}}{s}\right)\left\|\boldsymbol{x}_n^{(\kappa^{(r)}_n)}\right\|^2.
\end{aligned}
\end{equation}
\emph{where the expectation is taken over the randomness of SGD.}

\noindent \emph{Proof:} See Appendix \ref{lemma1}. \hfill$\square$ 

Based on Lemma 1, the convergence performance of the proposed AFL after $R$ rounds is given by the following.

\noindent\textbf{Theorem 1.} \emph{After $R$ rounds of training, the expected gradient norm of the global loss function is upper bounded by}
\begin{equation}
\begin{aligned}
&\mathbb{E}\left\|\nabla F(\boldsymbol{z}^{(R)})\right\|^2 \leq \frac{4\left(F(\boldsymbol{w}^{(0)})  - F(\boldsymbol{w}^{*})\right)}{\eta R}
\\&+ \frac{4 L^2}{NR}\sum_{r=0}^{R-1}\sum_{n\in \mathcal{N}}\zeta_n^{(r)} \theta_n^{(r)}\left(5-\frac{3k_n^{(r)}}{s}\right)\left\|\boldsymbol{x}_n^{(r)}\right\|^2 \\
&+\frac{8 \eta^2 L^2 G^2}{NR}\sum_{r=0}^{R-1}\sum_{n\in \mathcal{N}}\left(\theta_n^{(r)}\right)^2+\frac{4\eta L\sigma}{N}.
\label{theorem11}
\end{aligned}
\end{equation}
\emph{where $\boldsymbol{z}^{(R)}$ is a random variable sampling a previous parameter $\boldsymbol{w}_n^{(r)}$ with probability $\frac{1}{NR}$, and the expectation is taken over the randomness of SGD.}

\noindent \emph{Proof:} See Appendix \ref{theorem1}. \hfill$\square$ 

\noindent\textbf{Remark 1.} Theorem 1 indicates that, when all other variables are fixed, the AFL convergence primarily depends on the sparsification degree $k_n^{(r)}$ and model staleness $\theta_n^{(r)}$. A higher $k_n^{(r)}$ reduces the right-hand side (RHS) of the bound in (\ref{theorem11}), accelerating convergence, while a larger $\theta_n^{(r)}$ increases the RHS, slowing it down. Furthermore, the effects of the term $\left(5 - \frac{3k_n^{(r)}}{s}\right)$ and $\theta_n^{(r)}$ are intertwined: when one is large, it amplifies the negative impact of the other. Sparse gradients are more sensitive to model staleness, as outdated and sparse gradients contribute less effectively to model improvement. Likewise, high staleness exacerbates the negative impact of sparsification. Thus, convergence is significantly degraded when both sparsification error and model staleness are large.

Mobility affects the contact and inter-contact times, which in turn impact sparsification and model staleness. To quantify this effect, we analyze how contact time, inter-contact time, and device speed impact AFL convergence. For analytical simplicity, we assume each device $n$ maintains a constant transmission rate $A_n$ across all rounds. We analyze the impact of contact and inter-contact times on model staleness and present Lemma 2.

\noindent\textbf{Lemma 2.} \emph{The second moment of model staleness of every device $n\in\mathcal{N}$ is bounded by}
\begin{equation}
\begin{aligned}
&\mathbb{E}\left[\left(\theta^{(r)}_n\right)^2\right] \leq 1+\frac{\lambda_n}{\lambda_n+c_n} \cdot \frac{e^{-\frac{4\delta}{\lambda_n}}-3e^{{-\frac{3\delta}{\lambda_n}}}+4e^{{-\frac{2\delta}{\lambda_n}}}}{1-2e^{{-\frac{\delta}{\lambda_n}}}+e^{{-\frac{2\delta}{\lambda_n}}}},
\label{aggint}
\end{aligned}
\end{equation}
\emph{where the expectation is taken over the randomness of contact and inter-contact times.}

\noindent \emph{Proof:} See Appendix \ref{lemma2}. \hfill$\square$ 

For notational convenience, we denote the right-hand side of (\ref{aggint}) as $\Theta_n$ in the following. Then, we analyze the impact of contact time and on sparsification error, obtaining Lemma 3.

\noindent \textbf{Lemma 3.} \emph{The sparsification error is upper bounded by}
\begin{equation}\mathbb{E} \|\boldsymbol{x}^{(r)}_n-S_n^{(r)}(\boldsymbol{x}^{(r)}_n)\|^2\leq\left(1- e^{-\frac {(u+\log_2s)}{A_nc_n}} \right)\|\boldsymbol{x}^{(r)}_n\|^2,\label{spar}
\end{equation}
\emph{where the expectation is taken over the randomness of contact time.}

\noindent \emph{Proof:} See Appendix \ref{lemma3}. \hfill$\square$ 

Let $\gamma_n=e^{-\frac {(u+\log_2s)}{A_n c_n}}$ in the following. Combining Lemmas 2, 3 and Theorem 1, we have Theorem 2.

\noindent\textbf{Theorem 2.} \emph{After $R$ rounds of training, the expected gradient norm of the global loss function is upper bounded by}
\begin{equation}
\begin{aligned}
&\mathbb{E}\left\|\nabla F(\boldsymbol{z}^{(R)})\right\|^2 \leq \frac{8L\left(F(\boldsymbol{w}^{(0)})  - F(\boldsymbol{w}^{*})\right)}{\sqrt{ R}}+\frac{2\sigma}{N\sqrt{R}}
\\& +\frac{ G^2}{NR}\sum_{n\in \mathcal{N}} \frac{(16-8
\gamma_n-11\gamma_n^2+6\gamma_n^3)\Theta_n}{\gamma_n^2}.
\label{theorem12}
\end{aligned}
\end{equation}
\emph{where the expectation is taken over the randomness of contact time, inter-contact time and SGD.}

\noindent\emph{Proof:} See Appendix \ref{theorem2}. \hfill$\square$

\noindent\textbf{Remark 2.} Theorem 2 shows that, when other parameters are fixed, the convergence bound increases with $\Theta_n$. Taking the partial derivative of the RHS of (\ref{theorem12}) with respect to $\gamma_n$, we get $ \frac{(6\gamma_n^3+8\gamma_n - 32)\Theta_n G^2}{\gamma_n^3NR}$, which is always negative for $0\leq \gamma_n \leq 1$. Thus, the convergence bound decreases with $\gamma_n$. From (\ref{aggint}), $\Theta_n$ increases with average contact time $c_n$ and decreases with average inter-contact time $\lambda_n$, while (\ref{spar}) shows that $\gamma_n$ decreases with $c_n$. This implies that increasing $c_n$ improves convergence, as both $\gamma_n$ and $\Theta_n$ lower the bound, while increasing $\lambda_n$ degrades convergence due to a rise in $\Theta_n$.

To further investigate the impact of device speed on convergence performance, we simplify the system model by assuming that all devices share the same average speed $v$ and transmission rate $A$. In a wide range of mobility models, such as random waypoint, random walk, and random direction models, it is shown that both average contact time and inter-contact time are inversely proportional to the average device speed \cite{ispeed1,ispeed3}. Specifically, $c_n = C/v$ and $\lambda_n = \Lambda/v$, where $v$ is the average device speed. $C$ and $\Lambda$
are constants dependent on the communication range and mobility area of devices. Under these assumptions, and based on the fact that $0\leq \gamma_n \leq 1$ and $\Theta_n \geq 1$, we derive Corollary 1.

\noindent\textbf{Corollary 1.} \emph{Assuming that the contact time and inter-contact time are inversely proportional to device speeds, the expected gradient norm of the global loss function is bounded by}
\begin{equation}
\begin{aligned}
&\mathbb{E}\left\|\nabla F(\boldsymbol{z}^{(R)})\right\|^2 \leq \frac{8L\left(F(\boldsymbol{w}^{(0)})- F(\boldsymbol{w}^{*})\right)}{\sqrt{R}}+ \frac{2\sigma}{N\sqrt{R}}+\\&\frac{16G^2 e^{\frac{2(u+\log_2s)v}{AC}}}{R} \left(1+\frac{\Lambda}{\Lambda+C} \frac{e^{-\frac{4\delta v}{\Lambda}}-3e^{-\frac{3\delta v}{\Lambda}}+4e^{-\frac{2\delta v}{\Lambda}}}{e^{-\frac{2\delta v}{\Lambda}}-2e^{-\frac{\delta v}{\Lambda}}+1}\right). \label{conspeed}
\end{aligned}
\end{equation}

\noindent\textbf{Remark 3.} As $v$ increases from $0$ to infinity, the convergence bound in Corollary 1 first decreases and then increases. This implies that mobility improves FL convergence at lower speeds but degrades it at higher speeds. At lower speeds, devices have sufficient contact time to upload the gradients, and the negative effect of sparsification error is marginal. Moderate increases in device speed increase the contact rate, enhancing synchronization with the global model and reducing model staleness, which in turn improves FL convergence. However, at higher speeds, severely limited contact time restricts gradient transmission, and the effect of sparsification error becomes increasingly pronounced, leading to a decline in performance.

\section{Dynamic Sparsification Control Algorithm}
Convergence analysis shows that AFL convergence is mainly influenced by the sparsification degree and model staleness. Although a higher sparsification degree mitigates sparsification error, it requires uploading more gradients, and increases energy consumption for transmissions. This issue is particularly critical in practical AFL scenarios where mobile devices, such as vehicles or UAVs operating at the wireless edge, are constrained by limited energy resources. Since the energy supply is often non-replenishable during training, efficient energy usage is essential.

As shown in Theorem 1, the impacts of sparsification and model staleness on convergence are mutually reinforcing. When staleness is high, sparse and outdated gradients contribute little to model improvement, necessitating higher sparsification to preserve update effectiveness. This highlights the need to adapt the sparsification level in response to model staleness to maintain training efficiency. To this end, we propose the MADS algorithm, which adaptively adjusts the sparsification degree and transmission power to enhance the AFL convergence while ensuring long-term energy efficiency.

\subsection{Problem Formulation}
To enable practical algorithm design, we focus on Theorem 1, which provides a round-wise bound suitable for online optimization. We minimize the RHS of (\ref{theorem11}) by optimizing the sparsification degree $k_n^{(r)}$ and transmission power $p_n^{(r)}$. This is equivalent to minimizing $\sum_{r=0}^{R-1}\sum_{n\in \mathcal{N} }\zeta_n^{(r)} \theta_n^{(r)}\left(5-\frac{3k_n^{(r)}}{s}\right)\left\|\boldsymbol{x}_n^{(r)}\right\|^2$, as all other terms are constant. We denote $U_n^{(r)}=\zeta_n^{(r)} \theta_n^{(r)}\left(5-\frac{3k_n^{(r)}}{s}\right)\left\|\boldsymbol{x}_n^{(r)}\right\|^2$, and formulate the optimization problem as follows.
\begin{subequations}
\begin{align}
&P0:  \min_{\boldsymbol{k}, \boldsymbol{p}} \sum_{r=0}^{R-1}\sum_{n\in \mathcal{N}}U_n^{(r)} \label{obj0} \\
& \text{s.t.} \   \sum_{r=0}^{R-1} E^{(r)}_n \leq E^{\text{con}}_n,\quad \forall n \in \mathcal{N},  \label{energy1}\\
& \quad \ \ uk_{n}^{(r)}+k_{n}^{(r)}\log_2 s \leq \tau^{(r)}_n A_{n}^{(r)}, \  \forall r \in \mathcal{R}, \ \forall n \in \mathcal{N},\label{transmission1}\\
& \quad \ \ 0\leq k_{n}^{(r)} \leq s, \quad \forall r \in \mathcal{R}, \ \forall n \in \mathcal{N},\label{k}\\
& \quad \ \ 0\leq p_n^{(r)} \leq p^{\text{max}}_n , \quad \forall r \in \mathcal{R}, \ \forall n \in \mathcal{N},\label{p}
\end{align}
\end{subequations}
where $\boldsymbol{k}^{(r)} = [k_{1}^{(r)}, k_{2}^{(r)},\ldots,k_{N}^{(r)}]$ is the sparsification degree, and $\boldsymbol{k} = [\boldsymbol{k}^{(1)}, \boldsymbol{k}^{(2)}, \ldots, \boldsymbol{k}^{(R)}]$. $\boldsymbol{p}^{(r)} = [p_{1}^{(r)}, p_{2}^{(r)},\ldots,p_{N}^{(r)}]$ denotes the power allocation, and $\boldsymbol{p} = [\boldsymbol{p}^{(1)}, \boldsymbol{p}^{(2)}, \ldots, \boldsymbol{p}^{(R)}]$. Constraint (\ref{energy1}) ensures the total energy consumption of device $n$ does not exceed its energy budget. Constraint (\ref{transmission1}) ensures that devices finish transmission within the contact time. Constraints (\ref{k}) and  (\ref{p}) limit the range of optimization variables. 

The transmission rate of device $n$ in round $r$ is given by
$$A_{n}^{(r)}=B_n \log_2 \left(1+\frac{p_n^{(r)}|h_n^{(r)}|^2}{B_n N_0}\right),$$
where $B_n$ denotes the allocated bandwidth, $h_n^{(r)}$ is the channel coefficient, and $N_0$ is the noise power spectral density. The energy consumed by device $n$ in round $r$ is modeled as
$$E_n^{(r)}=\frac{p_n^{(r)} \left(uk_{n}^{(r)}+k_{n}^{(r)}\log_2 s \right)} {A_{n}^{(r)}}.$$

\subsection{Problem Transformation and Solution}
The original problem $P0$ is a \emph{stochastic optimization problem} due to the unknown system states, such as channel conditions and contact times. Solving it offline requires full knowledge of future system states, which is infeasible in practice. We therefore reformulate it as an online problem using the drift-plus-penalty framework from Lyapunov optimization \cite{stochastic}, enabling decisions based on current system states. To handle the long-term energy constraint, we introduce a virtual energy queue $\boldsymbol{q}^{(r)} = [q_1^{(r)}, \dots, q_{N}^{(r)}]$, where each $q_n^{(r)}$ tracks the deviation of device $n$'s cumulative energy consumption from its average per-round energy budget, evolving as:
\begin{equation}
q_{n}^{(r+1)} =\max\left\{q_{n}^{(r)}+ E^{(r)}_n-\frac{E_n^{\text{con}}}{R},0\right\}.\label{queue1}
\end{equation}
All virtual queues are initialized to 0, i.e., $\boldsymbol{q}^{(0)}=\boldsymbol{0}$. Based on this queue structure, we reformulate the original problem $P0$ into the following online optimization problem $P1$.
\begin{subequations}
\begin{align}
&P1:  \min_{\boldsymbol{k}^{(r)}, \boldsymbol{p}^{(r)}} \sum_{n \in \mathcal{N}} V
U_n^{(r)}
  +  E^{(r)}_n q_{n}^{(r)}  \label{obj1} \\
&\text{s.t.} \  uk_{n}^{(r)}+k_{n}^{(r)}\log_2 s \leq \tau^{(r)}_n A_{n}^{(r)}, \quad \forall n \in \mathcal{N},\\
&\quad \ \ 0\leq k_{n}^{(r)} \leq s, \quad \forall n \in \mathcal{N},\\
&\quad \ \ 0\leq p_n^{(r)} \leq p^{\text{max}}_n , \quad \forall n \in \mathcal{N},
\end{align}
\end{subequations}
where $V$ is a weight parameter. The following theorem establishes a performance bound for the online solution of problem $P1$, in comparison to the optimal offline solution of problem $P0$. Superscript $^\dagger$ denotes the solution to $P1$, and $^*$ denotes the optimal offline solution to $P0$.

\noindent \textbf{Theorem 3.} \emph{Suppose all queues are initialized to zero. Then, the gap between the optimal value of $P0$ and that of the online solution to $P1$ is bounded by:}
\begin{equation}
\sum_{r=0}^{R-1}\sum_{n \in \mathcal{N}} 
 U_{n}^{*(r)} -  
 U_{n}^{\dagger(r)} \leq \frac{R^2 \Phi}{V},
\end{equation}
\emph{the total energy consumption of device $n$ is bounded by:}
\begin{equation}
\sum_{r=0}^{R-1} E^{(r)}_n \leq E^{\text{con}}_n+ \sqrt{2R^2 \Phi -2\sum_{r=0}^{R-1}\sum_{n\in\mathcal{N}}V U_{n}^{*(r)}},
\end{equation}
\emph{where $\epsilon_n^{(r)} \triangleq  E^{(r)}_n- \frac{E^{\text{con}}_n}{R}$, $\phi_n \triangleq \max_r\{|\epsilon_n^{(r)}|\}$, and $\Phi \triangleq \sum_{n\in \mathcal{N}}(\phi_n)^2$.}

\emph{Proof:} See Appendix \ref{lyp}.

Theorem 3 shows that, instead of directly solving the long-term stochastic optimization problem $P0$, we can alternatively solve the online problem $P1$. This approach guarantees bounded performance relative to the optimal offline solution of $P0$, while ensuring that the energy consumption remains within acceptable limits. The trade-off between the convergence performance and the energy consumption is balanced by the weight parameter $V$.

In $P1$, the optimization variables associated with different devices are decoupled in both the objective function and the constraints. Therefore, the problem $P1$ is decomposed into $N$ independent subproblems, one for each device:
\begin{subequations}
\begin{align}
& P2:  \min_{ k_{n}^{(r)}, p_{n}^{(r)}} V
\zeta_n^{(r)} \theta_n^{(r)} \left(5-\frac{3k_n^{(r)}}{s}\right)\left\|\boldsymbol{x}_n^{(r)}\right\|^2 +  E^{(r)}_n q_{n}^{(r)} \label{objp2} \\
&\text{s.t.} \ uk_{n}^{(r)}+k_{n}^{(r)}\log_2 s \leq \tau^{(r)}_n A_{n}^{(r)},\label{trans2} \\
&\quad \ \ 0\leq k_{n}^{(r)} \leq s,\label{spar2}\\
&\quad \ \ 0\leq p_n^{(r)} \leq p^{\text{max}}_n.\label{power2}
\end{align}
\end{subequations}
$P2$ is not a convex optimization problem, since the objective function (\ref{objp2}) is not convex. Exploring the problem structure, we have the following proposition.

\noindent \textbf{Proposition 1.} \emph{For the optimal solution to $P2$, constraint (\ref{trans2}) must be tight, i.e.,} 
\begin{equation}k_{n}^{(r)} = \frac{\tau^{(r)}_n A_{n}^{(r)}}{u + \log_2s}.\label{tight}\end{equation}
\noindent \emph{Proof:} This proposition is proved by contradiction. Assume that there exists an optimal solution $\{k_{n}^{*(r)}, p_{n}^{*(r)} \}$ such that the constraint (\ref{trans2}) is not tight, i.e., $uk_{n}^{*(r)}+k_{n}^{*(r)}\log_2 s < \tau^{(r)}_n A_{n}^{*(r)}.$ Consider another solution $\{k_{n}^{'(r)}, p_{n}^{'(r)} \}$, where $k_{n}^{'(r)} = k_{n}^{*(r)}$ and the constraint (\ref{trans2}) is tight. Then, we have $0\leq p_{n}^{'(r)} < p_{n}^{*(r)}$ and $E^{'(r)}_n < E^{*(r)}_n$. This alternative solution is feasible, and the corresponding objective value satisfies:
\begin{equation}
\begin{aligned}&
V
\zeta_n^{(r)} \theta_n^{(r)} \left(5-\frac{3k_n^{'(r)}}{s}\right)\left\|\boldsymbol{x}_n^{(r)}\right\|^2 +  E^{'(r)}_n q_{n}^{(r)}
 \\&< 
V
\zeta_n^{(r)} \theta_n^{(r)} \left(5-\frac{3k_n^{*(r)}}{s}\right)\left\|\boldsymbol{x}_n^{(r)}\right\|^2 +  E^{*(r)}_n q_{n}^{(r)}.\notag\end{aligned}
\end{equation}
This contradicts the assumption that $\{k_{n}^{*(r)}, p_{n}^{*(r)} \}$ is optimal. Proposition 1 is proved. \hfill$\square$

Based on Proposition 1, we set $k_{n}^{(r)} = \frac{\tau^{(r)}_n A_{n}^{(r)}}{u + \log_2s}$, and reduce $P2$ to $P3$.
\begin{subequations}
\begin{align}
&P3:  \min_{ p_{n}^{(r)}} - \frac{ 3 V \zeta_n^{(r)} \theta_n^{(r)} 
 \tau^{(r)}_n B_n \left\|\boldsymbol{x}_n^{(r)}\right\|^2}{s \left(u + \log_2s \right)}  \log_2 \left(1+\frac{p_n^{(r)}|h_n^{(r)}|^2}{B_n N_0}\right)  \notag \\&
  \quad \quad \quad \ \  +5V
\zeta_n^{(r)} \theta_n^{(r)} \left\|\boldsymbol{x}_n^{(r)}\right\|^2 + \tau^{(r)}_n p_{n}^{(r)} q_{n}^{(r)}   \label{obj4} \\
&\text{s.t.} \ 
0\leq p_{n}^{(r)} \leq \frac{B_n N_0}{|h_n^{(r)}|^2}\left(2^ {\frac{s \left(u + \log_2s \right)}{\tau^{(r)}_n B_n}}-1 \right),\\
&\quad \ \ 0\leq p_{n}^{(r)} \leq p_{n}^{\text{max}}.
\end{align}
\end{subequations}

\begin{algorithm}[!t]	
    \caption{The MADS algorithm} 
    \label{Algo2}
	\begin{algorithmic}
        \STATE{\textbf{Output:} The solution $\boldsymbol{k}^*, \boldsymbol{p}^*$ to $P0$;}
        \STATE{Initialize $\boldsymbol{k}^*=\boldsymbol{0}$ and $\boldsymbol{p}^*=\boldsymbol{0}$;}
        \FOR{$r$ \textbf{in} $\mathcal{R}$}
	\FOR{$n$ \textbf{in} $\mathcal{N}$}
        \IF{$\zeta_{n}^{(r)} = 1$}
        \STATE{Solve $P3$ based on Propositions 1 and 2 to obtain $k_{n}^{*(r)} $and $p_{n}^{*(r)}$;}
        \ELSIF{$\zeta_{n}^{(r)} = 0$}
        \STATE{Set $k_{n}^{*(r)}=p_{n}^{*(r)}=0$;}
        \ENDIF
	\ENDFOR
        \STATE{Update $\boldsymbol{q}^{(r)}$ according to (\ref{queue1}).}
        \ENDFOR
	\end{algorithmic}
\end{algorithm}

$P3$ is a convex optimization problem, since the objective function and all constraints are convex. The optimal solution is derived using the Karush-Kuhn-Tucker (KKT) conditions, as stated in Proposition 2.

\begin{figure*}[!t]
  \centering
  \begin{minipage}[t]{0.329\textwidth}
    \vspace{0pt}
    \centering
    \includegraphics[width=\textwidth]{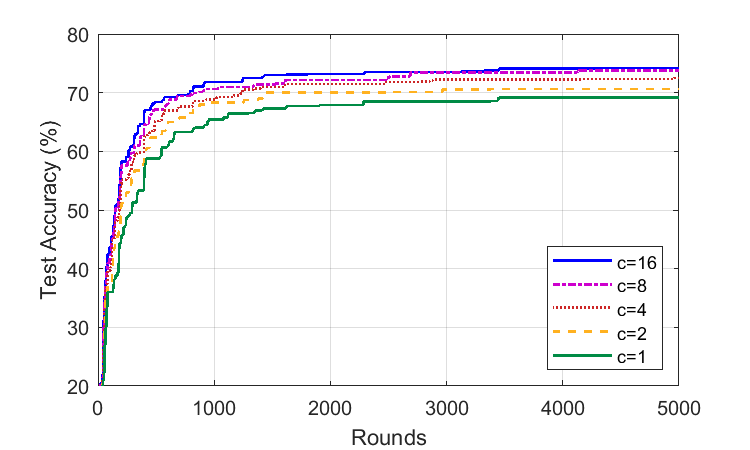}
    \caption{Test accuracy w.r.t. training rounds under different contact time (s).}
    \label{contact1}
  \end{minipage}
    \begin{minipage}[t]{0.329\textwidth}
    \vspace{0pt}
    \centering
    \includegraphics[width=\textwidth]{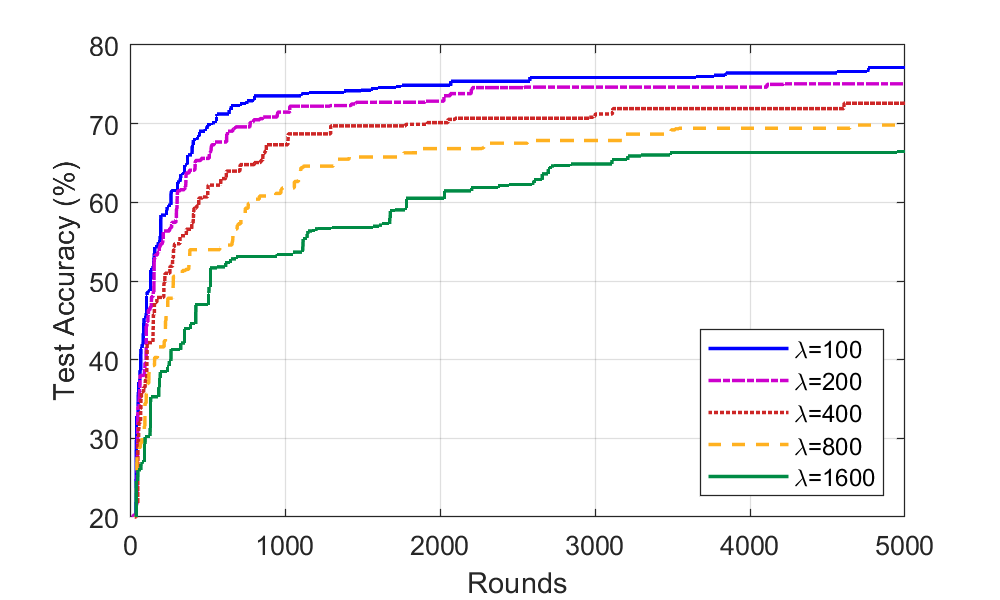}
    \caption{Test accuracy w.r.t. training rounds under different inter-contact time (s).}
    \label{intercontact1}
  \end{minipage}
  \begin{minipage}[t]{0.329\textwidth}
    \vspace{0pt} 
    \centering
    \includegraphics[width=\textwidth]{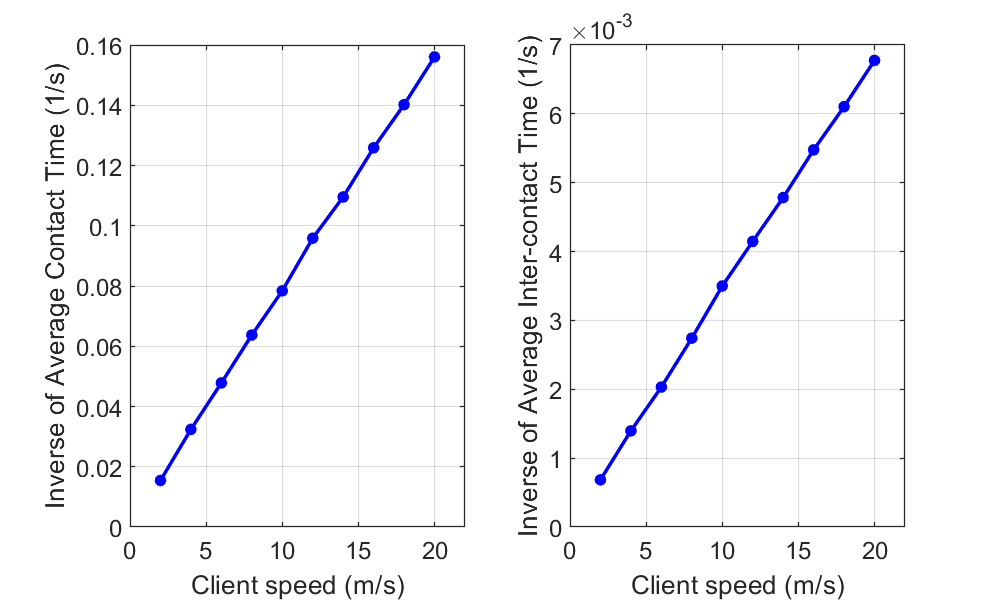}
    \caption{Effect of device speeds on contact time and inter-contact time.}
    \label{speedtic}
  \end{minipage}
\end{figure*}

\begin{figure*}[!t]
  \centering
  \begin{minipage}[t]{0.329\textwidth}
    \vspace{0pt}
    \centering
    \includegraphics[width=\textwidth]{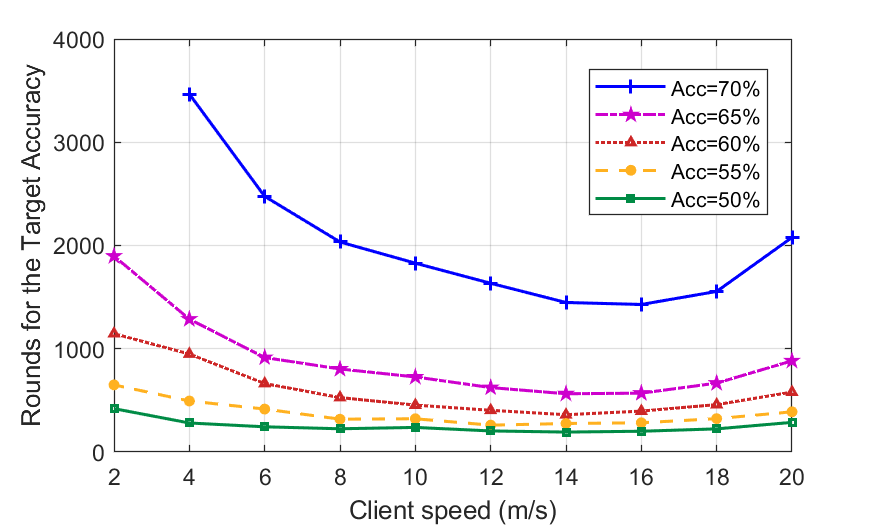}
    \caption{Test accuracy under different device speeds.}
    \label{velocity}
  \end{minipage}
    \begin{minipage}[t]{0.329\textwidth}
    \vspace{0pt}
    \centering
    \includegraphics[width=\textwidth]{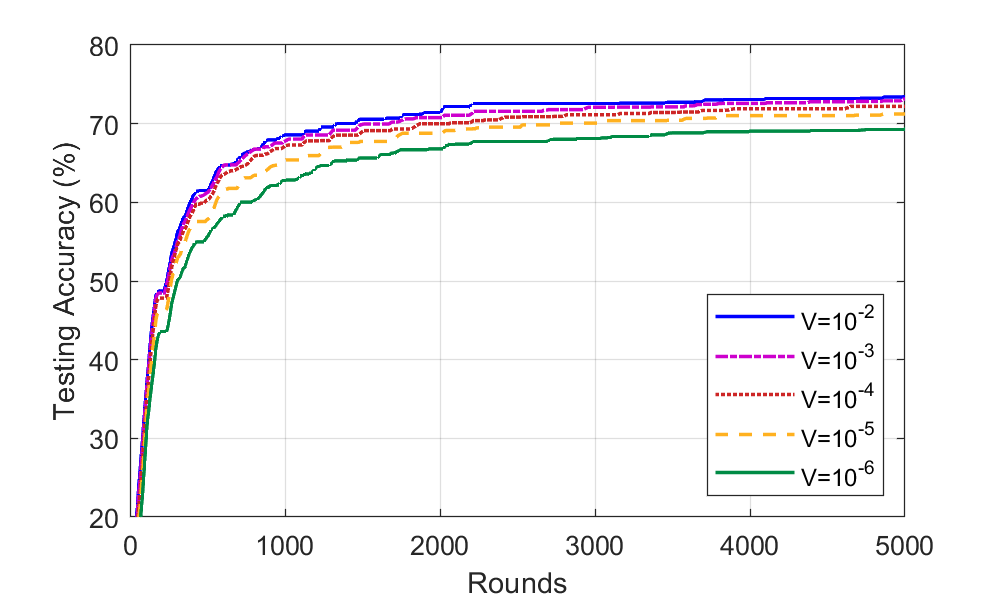}
    \caption{Test accuracy under different weight $V$.}
    \label{V}
  \end{minipage}
  \begin{minipage}[t]{0.329\textwidth}
    \vspace{0pt} 
    \centering
    \includegraphics[width=\textwidth]{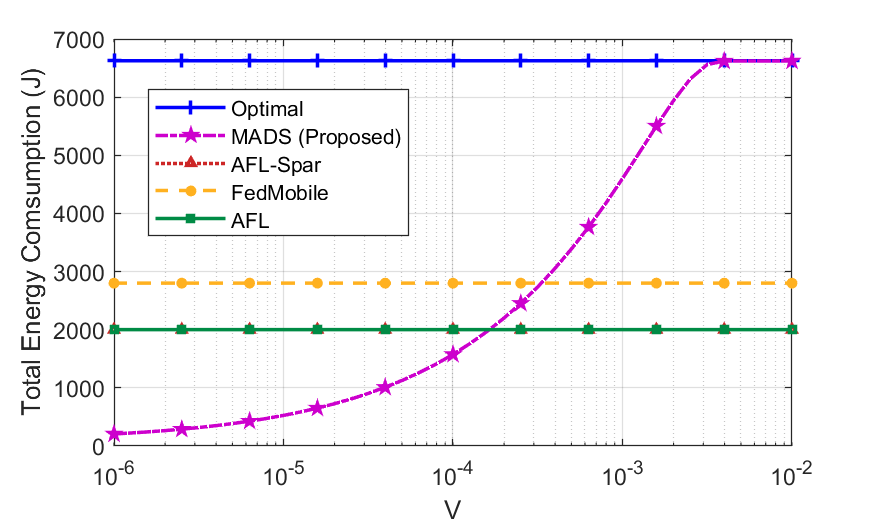}
    \caption{Energy consumption under different $V$.}
    \label{engsta}
  \end{minipage}
\end{figure*}

\noindent \textbf{Proposition 2.} \emph{The optimal power allocation for P3 is} 
\begin{equation}
p_n^{*(r)} = \left[\frac{3 V \zeta_n^{(r)} \theta_n^{(r)} 
  B_n \left\|\boldsymbol{x}_n^{(r)}\right\|^2}{q_{n}^{(r)}s(u + \log_2s)  } - \frac{B_n N_0}{|h_n^{(r)}|^2}  \right]^{P^{(r)}_n}_0,
\end{equation}
\emph{where} $P^{(r)}_n = \min \left(p^{\text{max}}_n ,\frac{B_n N_0}{|h_n^{(r)}|^2}\left(2^ {\frac{s \left(u + \log_2s \right)}{\tau^{(r)}_n B_n}}-1 \right)\right)$ \emph{and} $[a]^{P^{(r)}_n}_0$ \emph{denotes} $\min(\max(a,0), P^{(r)}_n)$.

Taking $p_n^{*(r)}$ back to (\ref{tight}), we get $k_n^{*(r)}$. Also, Proposition 2 shows that if $\zeta_n^{(r)}=0$, $p_n^{(r)}$ and $k_n^{(r)}$ can be directly set to $0$. Together, Propositions 1 and 2 provide closed-form solutions for optimal sparsification degree and power allocation, where $k_n^{*(r)}$ increases with both contact time $\tau^{(r)}_n$ and model staleness $\theta_n^{(r)}$, and $p_n^{*(r)}$ increases with $\theta_n^{(r)}$. As Theorem 2 and Corollary 1 suggest, higher mobility reduces both contact and inter-contact times, where the latter helps mitigate model staleness. Accordingly, under low mobility, where contact and inter-contact times are both long, MADS increases the sparsification degree and transmission power to upload more gradients and accelerate convergence. In contrast, under high mobility, MADS reduces the sparsification degree to ensure reliable uploads within the limited contact time.

The MADS algorithm is summarized in Algorithm \ref{Algo2}. Since both $k_n^{(r)}$ and $p_n^{(r)}$ have closed-form expressions, the per-round computational complexity is $\mathcal{O}(N)$. The algorithm can be executed in a decentralized manner. When a device contacts the MES, it solves $P3$ and adjusts its sparsification and power accordingly.

\section{Simulation Results}

This section presents the simulation results. We use the CIFAR-10 dataset \cite{cifar10} to train a ResNet-9 model \cite{resnet} with one MES and $20$ devices. The ResNet-9 model consists of nine convolutional layers with batch normalization, ReLU activations, residual blocks, a global average pooling layer, and a final fully connected layer, totaling $6,568,650$ parameters. The learning rate is set to $0.01$, and the batch size is $32$.

The dataset comprises $50,000$ training images and $10,000$ test images across $10$ classes, and is divided into $20$ subsets, each assigned to a device. We use a Dirichlet distribution to simulate the non-independent and identically distributed (non-i.i.d.) data distribution among devices, where $Z_i = \frac{z_i}{\sum_{i=1}^{I}z_i}$ denotes the proportion of the $i$-th class in a local dataset, and $I$ is the total number of classes. The variable $z_i$ follows a Gamma distribution, i.e., $z_i \sim \Gamma(\rho\bar{Z}_i, 1)$, where $\bar{Z}_i$ denotes the class distribution of the full dataset, and $\rho$ is the concentration parameter that controls the non-i.i.d. level. Small values of $\rho$ result in greater non-i.i.d. level, while larger values lead to more uniform distributions. As $\rho \rightarrow 0$, each device contains data from only one class. As $\rho \rightarrow \infty$, the data becomes identically distributed across all devices.

For wireless communications, we adopt the channel in the UMi-Street Canyon scenario in TR 38.901 \cite{pathloss}. The pathloss of the LOS and the NLOS channels are given by $PL_\text{LOS} = 32.4+21\log _{10}{d}+20\log _{10}{\beta}$, and $PL_\text{NLOS} = 32.4+31.9\log _{10}{d}+20\log _{10}{\beta}$, respectively, 
where $d$ is the distance between device and the MES, and $\beta$ is the carrier frequency. The simulation parameters are shown in Table \ref{table}.

\subsection{Impact of Mobility on the AFL}
In this part, we evaluate the impact of average contact times, inter-contact times, and device speeds on the AFL convergence.
\subsubsection{Impact of contact time}
Firstly, we evaluate the AFL convergence under different average contact times. Fig. \ref{contact1} illustrates the test accuracy with respect to training rounds, where both the convergence rate and the final test accuracy increase as the average contact time increases from $1$s to $16$s. This is because, as the average contact time increases, devices can upload more gradients, reducing the sparsification error. This validates Theorem 2, since the RHS of (\ref{theorem12}) decreases as the average contact time increases.

\begin{table}[t!]
\caption{Simulation Parameters.}
\begin{tabular}{l|l}
\hline
\textbf{Simulation Parameters}                   & \textbf{Values}                            \\ \hline
Device Bandwidth             & $1$MHz                             \\ \hline
Carrier Frequency            & $3.5$GHz                            \\ \hline
Maximum Transmission Power   & $0.2$W                              \\ \hline
Noise Power Spectral Density & -174dBm/Hz                        \\ \hline
Shadowing Fading Std. Dev.   & $4$dB (LOS) / $8.2$dB (NLOS)      \\ \hline
Energy Constraints        & Randomly selected from $50$J to $150$J\\ \hline
Round duration        & $10$s\\ \hline
\end{tabular}
\label{table}
\end{table}

\subsubsection{Impact of inter-contact time}
Then, we evaluate the AFL convergence under varying average inter-contact times. Fig. \ref{intercontact1} illustrates the test accuracy with respect to training rounds, showing that both the convergence rate and final test accuracy improve as the average inter-contact time decreases from $1600$s to $100$s. This aligns with the theoretical analysis in (\ref{theorem12}). A shorter inter-contact time enables more frequent model aggregation, thereby reducing model staleness and enhancing synchronization between local and global models. 


\begin{figure*}[htbp]
  \centering
  \begin{subfigure}[b]{0.329\linewidth}
    \captionsetup{justification=centering}  
    \includegraphics[width=\linewidth]{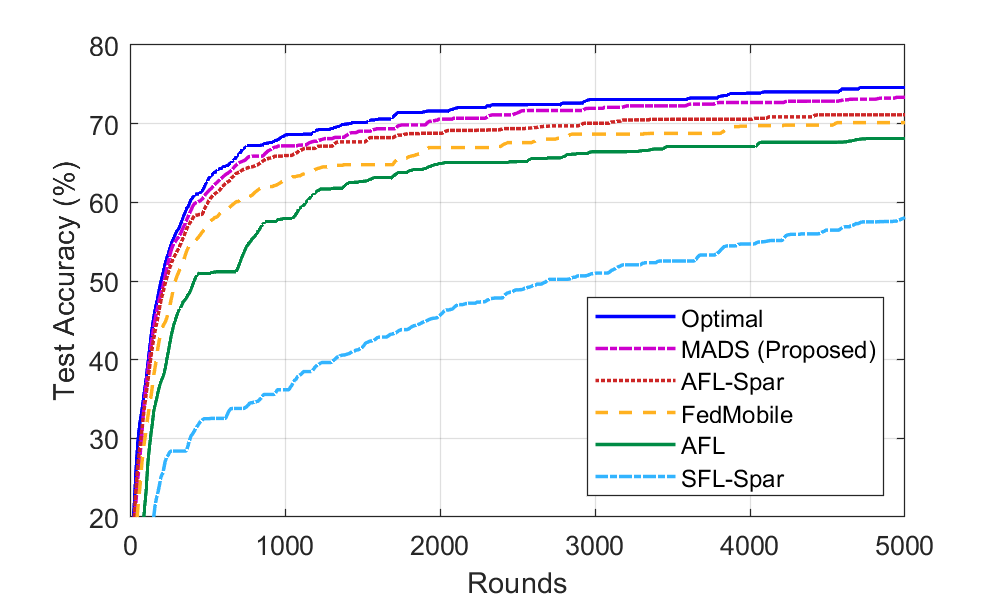}
    \caption{$\rho = 10$.}
    \label{noniid10}
  \end{subfigure}
  \begin{subfigure}[b]{0.329\linewidth}
    \captionsetup{justification=centering}  
    \includegraphics[width=\linewidth]{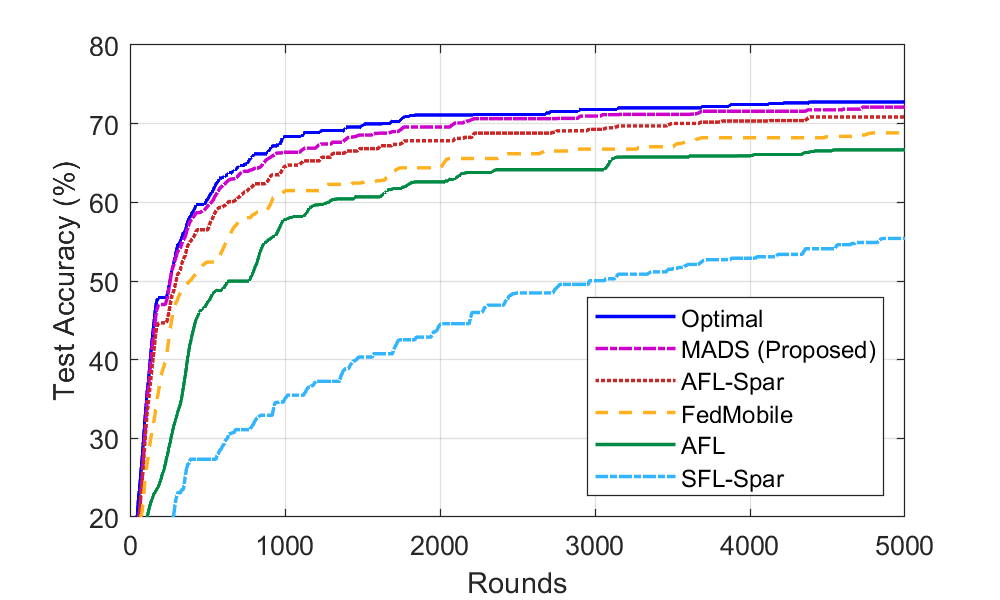}
    \caption{$\rho = 1$}
    \label{noniid1}
  \end{subfigure}
  \begin{subfigure}[b]{0.329\linewidth}
    \captionsetup{justification=centering}  
    \includegraphics[width=\linewidth]{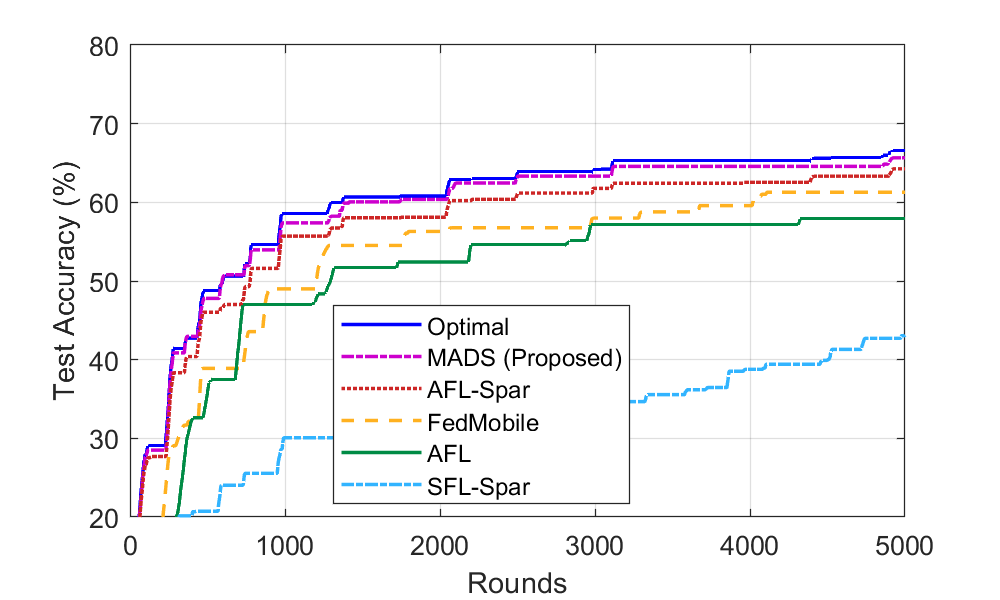}
    \caption{$\rho = 0.1$}
    \label{noniid01}
  \end{subfigure}
  \caption{Test accuracy of the proposed algorithm compared with the benchmarks under different levels of non-i.i.d. data.}
  \label{noniid}
\end{figure*}

\begin{figure*}[!t]
  \centering
  \begin{minipage}[t]{0.329\textwidth}
    \vspace{0pt}
    \centering
    \includegraphics[width=\textwidth]{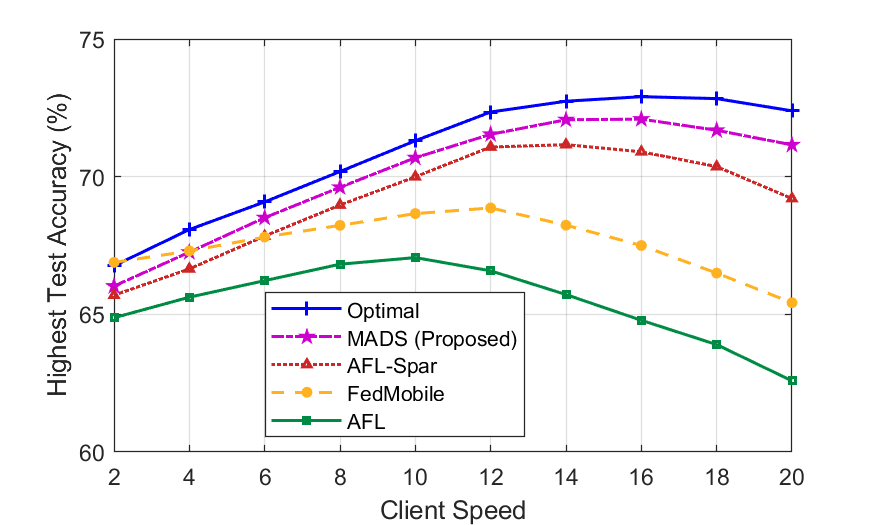}
    \caption{Test accuracy under different device speeds.}
    \label{stalcon}
  \end{minipage}
    \begin{minipage}[t]{0.329\textwidth}
    \vspace{0pt}
\centering
\includegraphics[width=\textwidth]{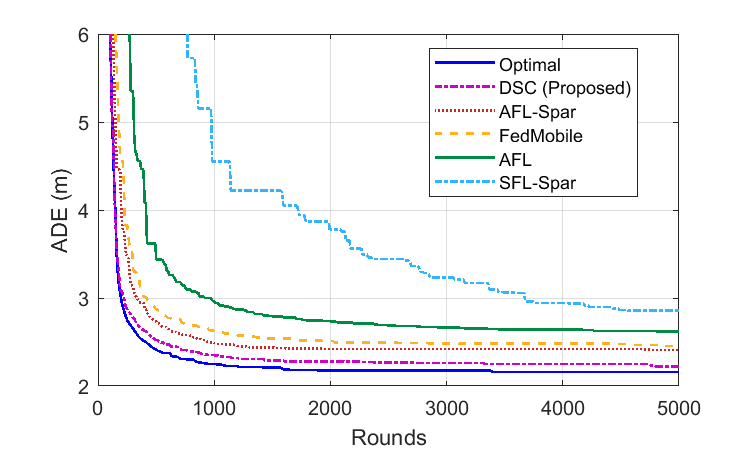}
\caption{ADE w.r.t. rounds on the Argoverse dataset.}    \label{adeiid}
  \end{minipage}
  \begin{minipage}[t]{0.329\textwidth}
    \vspace{0pt} 
\centering
\includegraphics[width=\textwidth]{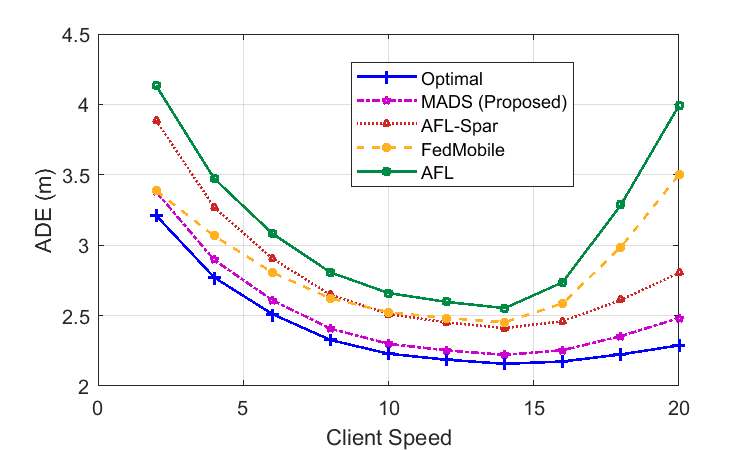}
\caption{ADE under different device speeds on the Argoverse dataset.}    \label{adeiid2}
  \end{minipage}
\end{figure*}
 
\subsubsection{Impact of device speeds}
We evaluate the impact of device speed on convergence performance using the random waypoint mobility model \cite{ispeed3}. In this model, each device randomly selects a destination within the simulation area and moves toward it at a speed uniformly distributed with mean $v$ m/s. Upon arrival, the device pauses for a random duration before selecting a new destination and repeating the process. In the simulation, the MES and $20$ devices move within a $1000$ m × $1000$ m area, with each device having a transmission range of $100$ m. As illustrated in Fig. \ref{speedtic}, an inversely proportional relationship is observed between device speed and both contact and inter-contact times, where higher device speeds result in lower contact and inter-contact times.

Fig.~\ref{velocity} shows the number of training rounds required to reach target accuracies under different device speeds. A smaller number of rounds indicates a faster convergence rate. The convergence rate initially increases and then decreases as device speed increases. This is because higher mobility reduces both the contact time and the inter-contact time, leading to a twofold effect on convergence performance. At lower speeds, the benefit of reduced inter-contact time outweighs the loss from shorter contact time, resulting in faster convergence. However, at higher speeds, the negative impact of reduced contact time dominates, leading to a slower convergence rate.

\subsection{Evaluation of the MADS Algorithm}

In this section, we evaluate the performance of the proposed MADS algorithm and compare it with several benchmarks, including: \emph{1) Synchronous FL with sparsification (SFL-Spar):} Devices update and upload their gradients when they contact the MES. In the inter-contact period, the device does not update its local model. \emph{2) Typical AFL method FedAsync (AFL) \cite{asy3}:} A state-of-the-art
AFL method to handle arbitrary communication patterns. \emph{3) AFL with sparsification (AFL-Spar):} An extension of AFL that incorporates top-$k$ sparsification, as described in Algorithm~\ref{Algo1}. \emph{4) FedMobile \cite{mobasy}:} A mobility-aware AFL scheme that exploits relay opportunities introduced by device mobility to accelerate convergence. \emph{5) Optimal Benchmark:} An idealized setup where devices are unconstrained by energy and transmit at maximum power, providing optimal performance under the same wireless settings.

\subsubsection{Performance under different values of $V$}
We first examine the effect of the weight parameter $V$ on MADS performance. As shown in Fig. \ref{V}, larger $V$ leads to faster convergence and higher final accuracy, since more energy is allocated for gradient transmission. Fig.~\ref{engsta} illustrates the total energy consumption versus $V$. As $V$ increases, the energy consumption of MADS increases accordingly. When $V$ exceeds $10^{-4}$, the energy constraints---reflected by the constant consumption levels of AFL and AFL-Spar (approximately 2000 J)---are violated. As $V$ approaches $10^{-2}$, all devices transmit with maximum power, and the total energy consumption of MADS stabilizes at the level of the optimal benchmark (around 6700 J), where devices operate without any energy limitations. This highlights the importance of tuning $V$ to balance energy efficiency and convergence performance in practical systems.

\subsubsection{Performance on different non-i.i.d. settings}
Then, we evaluate the performance of the MADS algorithm and compare it with benchmarks under different levels of non-i.i.d. data, as shown in Fig. \ref{noniid}. Among all benchmarks, SFL performs the worst due to intermittent connectivity, which limits effective model aggregation. AFL-Spar outperforms standard AFL and FedMobile, as the latter, without sparsification, may fail to transmit the gradients within the limited contact time. This highlights the necessity of applying sparsification in mobile FL scenarios. The proposed MADS algorithm considers long-term energy constraints and dynamically adjusts the sparsification degree, leading to further performance improvements over AFL-Spar and achieving close performance to the optimal benchmark. When $\rho = 0.1$, MADS achieves a test accuracy of $65.66\%$, exceeding AFL-Spar, FedMobile, AFL and SFL-Spar over $2.17\%$, $7.13\%$,  $13.27\%$ and $52.25\%$ respectively.

\subsubsection{Performance under different device speeds} We evaluate the performance of MADS under varying device speeds, as shown in Fig.~\ref{stalcon}. At low speeds, all schemes exhibit similar performance, as devices have sufficient contact time to upload gradients, so sparsification is less impactful. FedMobile performs slightly better by leveraging relay opportunities to increase the contact rate. 

As speed increases, the performance gap between MADS and the benchmark methods becomes more pronounced. At 20 m/s, MADS achieves test a accuracy of $71.15\%$, outperforming AFL-Spar, FedMobile, and AFL by $2.82\%$, $8.76\%$, and $13.69\%$, respectively. This is due to severe contact time limitations under high mobility. AFL and FedMobile, lacking sparsification, often suffer transmission failures. AFL-Spar and MADS mitigate this via sparsification, while MADS further adapts the sparsification degree and power allocation based on energy and model staleness, leading to superior performance. SFL performs significantly worse than the other methods, with the highest test accuracy remaining below $60\%$ across all speeds. Therefore, its curve is omitted from the plot.

\subsection{Evaluation on Argoverse Trajectory Prediction Dataset}
Finally, we evaluate the proposed MADS algorithm on the real-world Argoverse trajectory prediction dataset \cite{Argoverse}, which contains over $300,000$ sequences collected in Pittsburgh and Miami at 10 Hz. Each sequence involves predicting vehicle positions over the next 3 seconds. The dataset includes $205,942$ training, $39,472$ validation, and $78,143$ test sequences, uniformly divided into 20 subsets.

Based on the dataset, devices collaboratively train a LaneGCN model \cite{lane}, comprising ActorNet, MapNet, and FusionNet. The ActorNet uses a 1D convolutional neural network and a Feature Pyramid Network to extract trajectory features, MapNet applies GCNs to capture map features, and FusionNet fuses both for final predictions. The batch size is set to $16$, and the learning rate is $0.1$. We use ADE as the evaluation metric, defined as the mean Euclidean distance between predicted and ground-truth vehicle positions along the trajectory. Fig. \ref{adeiid} shows the ADE w.r.t. rounds, where the proposed MADS outperforms the benchmarks both in terms of convergence rate and final ADE. Fig. \ref{adeiid2} shows the ADE under different device speeds. The overall trend is that as the client speed increases, the ADE first increases and then decreases, consistent with the pattern observed on the CIFAR-10 dataset. The optimal benchmark consistently achieves the lowest ADE, followed closely by the proposed MADS algorithm. These results validate the effectiveness of the proposed AFL framework and MADS algorithm in real-world autonomous driving datasets.

\section{Conclusions}
This paper has investigated an AFL framework over mobile networks, where asynchronous aggregation and sparsification techniques are integrated to accommodate intermittent device connectivity. We have provided a convergence analysis to quantify the interplay among sparsification, model staleness and mobility-induced contact patterns, and their joint impact on AFL convergence. A MADS algorithm has been proposed to further enhance the AFL convergence by controlling the sparsification degree and power consumption. The convergence performance of the proposed AFL has been evaluated under varying contact times, inter-contact times and device speeds, validating the theoretical findings with experimental results. Furthermore, results have shown that under a high non-i.i.d. setting, the proposed MADS algorithm increases the image classification accuracy on the CIFAR-10 dataset by $8.76\%$ and reduces the ADE on the Argoverse trajectory prediction by dataset $9.46\%$ compared to the state-of-the-art benchmark.

\appendices
\section{Proof of Lemma 1} 
\label{lemma1}

Based on the Cauchy-Schwarz inequality, there is
\begin{equation}
\begin{aligned}
&\mathbb{E}\left\|\boldsymbol{v}^{(r)}-\boldsymbol{w}_{n}^{(r)}\right\|^{2}=4\mathbb{E}\left\|\boldsymbol{v}^{(r)}-\frac1N\sum_{m\in \mathcal{N}}\boldsymbol{v}^{(\kappa_m^{(r)})}\right\|^2\\&+4\mathbb{E}\left\|\frac1N\sum_{m\in \mathcal{N}}\boldsymbol{v}^{(\kappa_m^{(r)})}-\boldsymbol{w}^{(r)}\right\|^2\\&+4\mathbb{E}\left\|\boldsymbol{w}^{(r)}-\boldsymbol{w}^{(\kappa_n^{(r)})}\right\|^2+4\mathbb{E}\left\|\boldsymbol{w}^{(\kappa_n^{(r)})}-\boldsymbol{w}_n^{(r)}\right\|^2.\notag
\end{aligned}
\end{equation}

For the first term, there is
\begin{equation}
\begin{aligned}
&\mathbb{E}\left\|\boldsymbol{v}^{(r)}-\frac{1}{N}\sum_{m\in\mathcal{N}}\boldsymbol{v}^{(\kappa_{m}^{(r)})}\right\|^{2} \\
&\leq\frac{\eta^2}N\sum_{m\in\mathcal{N}}\left(r-\kappa_m^{(r)}\right)\sum_{j=\kappa_m^{(r)}}^{r-1}\mathbb{E}\left\|\nabla f_m\left(\boldsymbol{w}_m^{(j)},\mathcal{B}_m^{(j)}\right)\right\|^2 \\
&\leq\frac{1}{N}\sum_{m\in\mathcal{N}}\left(\theta_{m}^{(r)}\right)^{2}\eta^{2}G^2.\label{l21}\notag
\end{aligned}
\end{equation}

For the second term, there is
\begin{equation}
\begin{aligned}
&\mathbb{E}\left\|\frac1N\sum_{m\in\mathcal{N}}\boldsymbol{v}^{(\kappa_m^{(r)})}-\boldsymbol{w}^{(r)}\right\|^2 \\
&\leq\mathbb{E}\left\|\frac1N\sum_{m\in\mathcal{N}}\sum_{j=1}^{\kappa_m^{(r)}}\nabla f_m\left(\boldsymbol{w}_m^{(j-1)},\mathcal{B}_m^{(j-1)}\right)-\zeta_m^{(j)}S(\boldsymbol{x}_m^{(j)})\right\|^2\notag\\
&\leq\frac1N\sum_{m\in\mathcal{N}}\left\|\boldsymbol{e}_m^{(\kappa_m^{(r)})}\right\|^2 \overset{(a)}{\leq}\frac1N\sum_{m\in\mathcal{N}} (1-\frac{k_n^{(\kappa_m^{(r)})}}{s}) \left\|\boldsymbol{x}_m^{(\kappa_m^{(r)})}\right\|^2 , \label{l22}
\end{aligned}
\end{equation}
where $(a)$ holds according to the Lemma A.1 in \cite{spar1}.

For the third term, there is
\begin{equation}
\begin{aligned}
&\mathbb{E}\left\|\boldsymbol{w}^{(r)}-\boldsymbol{w}^{(\kappa_n^{(r)})}\right\|^2 \\&\leq\frac1N\sum_{m\in\mathcal{N}\setminus\{n\}}\left\|S(\boldsymbol{x}_m^{(\kappa_m^{(r)})})\right\|^2
\\&\leq \frac2N\sum_{m\in\mathcal{N}\setminus\{n\}}\left\|\boldsymbol{x}_m^{(\kappa_m^{(r)})}-S(\boldsymbol{x}_m^{(\kappa_m^{(r)})})\right\|^2+\left\|\boldsymbol{x}_m^{(\kappa_m^{(r)})}\right\|^2
\\&\leq \frac1N\sum_{m\in\mathcal{N}\setminus\{n\}} (4-\frac{2k_n^{(\kappa_m^{(r)})}}{s}) \left\|\boldsymbol{x}_m^{(\kappa_m^{(r)})}\right\|^2.\notag
\end{aligned}
\end{equation}

For the fourth term, there is
\begin{equation}
\begin{aligned}
&\mathbb{E}\left\|\boldsymbol{w}^{(\kappa_n^{(r)})}-\boldsymbol{w}_n^{(r)}\right\|^2 \leq\left(\theta_n^{(r)}\right)^2\eta^2G^2 .\notag
\end{aligned}
\end{equation}
Combining these terms, we prove Lemma 1, where the expectation is taken over the randomness of SGD.

\section{Proof of Theorem 1}
\label{theorem1}
 According to Assumption 1, the global loss function is also $L$-smooth. There is

\begin{align}
\label{t11}
&\mathbb{E}[F(\boldsymbol{v}^{(r)})]-F(\boldsymbol{v}^{(r-1)}) \\&\leq \mathbb{E}\langle\nabla F(\boldsymbol{v}^{(r-1)}),\boldsymbol{v}^{(r)}-\boldsymbol{v}^{(r-1)})\rangle+\frac L2 \mathbb{E}\|\boldsymbol{v}^{(r)}-\boldsymbol{v}^{(r-1)}\|^2. \notag 
\end{align}

For the first term, there is

\begin{align}
\label{t1f}
&\mathbb{E}\langle\nabla F(\boldsymbol{v}^{(r-1)}),\boldsymbol{v}^{(r)}-\boldsymbol{v}^{(r-1)})\rangle \notag
\\&=-\mathbb{E} \left\langle\nabla F(\boldsymbol{v}^{(r-1)}),\frac\eta N\sum_{n\in\mathcal{N}}\nabla F_n(\boldsymbol{w}^{(r-1)}_n)\right\rangle \notag
\\&=-\frac\eta2\|\nabla F(\boldsymbol{v}^{(r-1)})\|^2-\frac {\eta} {2N}\sum_{n\in \mathcal{N}}\left\|\nabla F_n(\boldsymbol{w}^{(r-1)}_n)\right\|^2 \notag
\\&+\frac{\eta}{2N}\sum_{n\in \mathcal{N}}\left\|\nabla F(\boldsymbol{v}^{(r-1)})-\nabla F_n(\boldsymbol{w}^{(r-1)}_n)\right\|^2 \notag
\\&\leq-\frac{\eta} {4N}\sum_{n\in\mathcal{N}}\mathbb{E}\left\|\nabla F(\boldsymbol{w}_n^{(r)})\right\|^2 -\frac {\eta} {2N}\sum_{n\in \mathcal{N}}\left\|\nabla F_n(\boldsymbol{w}^{(r-1)}_n)\right\|^2 \notag
\\&+\frac{\eta L^2}{N}\sum_{n\in \mathcal{N}}\left\|{\boldsymbol{v}}^{(r-1)}-\boldsymbol{w}^{(r-1)}_n\right\|^2.
\end{align}

For the second term in (\ref{t11}), there is

\begin{align}
&\frac L2 \mathbb{E}\|\boldsymbol{v}^{(r)}-\boldsymbol{v}^{(r-1)}\|^2 \notag
\\&\leq \frac L2 \mathbb{E} \left\|\frac\eta N\sum_{n\in\mathcal{N}}\nabla f_n(\boldsymbol{w}^{(r-1)}_n,\mathcal{B}^{(r-1)}_n)\right\|^2 \notag
\\ & \leq \frac{\eta^2L}N\sum_{n\in \mathcal{N}}\left\| \nabla f_n(\boldsymbol{w}^{(r-1)}_n,\mathcal{B}^{(r-1)}_n) - \nabla F_n(\boldsymbol{w}^{(r-1)}_n)\right\|^2 \notag
\\&+ \frac{\eta^2L}N\sum_{n\in \mathcal{N}}\left\|\nabla F_n(\boldsymbol{w}^{(r-1)}_n)\right\|^2\notag
\\
& \leq \frac{\eta^2L\sigma}{N}+\frac{\eta^2L}N\sum_{n\in \mathcal{N}}\left\|\nabla F_n(\boldsymbol{w}^{(r-1)}_n)\right\|^2.
\label{t1s}
\end{align}

Plugging (\ref{t1f}) and (\ref{t1s}) into (\ref{t11}), and setting $\eta \leq \frac{1}{2L}$ we have
\begin{equation}
\begin{aligned}
&\frac{\eta} {4N}\sum_{n\in\mathcal{N}}\mathbb{E}\left\|\nabla F(\boldsymbol{w}_n^{(r)})\right\|^2 \leq F(\boldsymbol{v}^{(r-1)})  - \mathbb{E}[F(\boldsymbol{v}^{(r)})]
\\& +\frac{\eta L^2}{N}\sum_{n\in \mathcal{N}}\left\|{\boldsymbol{v}}^{(r-1)}-\boldsymbol{w}^{(r-1)}_n\right\|^2+\frac{\eta^2L\sigma}{N}.
\notag
\end{aligned}
\end{equation}
Based on Lemma 1, and taking a telescopic sum from $r = 1$ to $r = R$, we get
\begin{equation}
\begin{aligned}
&\mathbb{E}\left\|\nabla F(\boldsymbol{z}^{(R)})\right\|^2 \leq \frac{4\left(F(\boldsymbol{w}^{(0)})  - F(\boldsymbol{w}^{*})\right)}{\eta R}
\\&+ \frac{4 L^2}{NR}\sum_{r=0}^{R-1}\sum_{n\in \mathcal{N}}\zeta_n^{(r)} \theta_n^{(r)}\left(5-\frac{3k_n^{(r)}}{s}\right)\left\|\boldsymbol{x}_n^{(r)}\right\|^2 \\
&+\frac{8 \eta^2 L^2 G^2}{NR}\sum_{r=0}^{R-1}\sum_{n\in \mathcal{N}}\left(\theta_n^{(r)}\right)^2+\frac{4\eta L\sigma}{N}.
\label{t12}\notag
\end{aligned}
\end{equation}
where $\boldsymbol{z}^{(R)}$ is sampled from $\{\boldsymbol{w}_n^{(r)}\}$ for $n \in \mathcal{N}$ and $r \in \mathcal{R}$ with probability $\Pr(\boldsymbol{z}^{(R)} = \boldsymbol{w}_n^{(r)}) = \frac{1}{RT}$, which implies $\mathbb{E}\left\|\nabla F(\boldsymbol{z}^{(R)})\right\|^2 = \frac{1} {NR}\sum_{r=0}^{R-1}\sum_{n\in\mathcal{N}}\mathbb{E}\left\|\nabla F(\boldsymbol{w}_n^{(r)})\right\|^2$. 
\section{Proof of Lemma 2}
\label{lemma2}
Ranking the elements of $\boldsymbol{x}^{(r)}_n$ in an ascending order, where $x_1\leq x_2 \leq...\leq x_{s}$, we have
\begin{equation}
\begin{aligned}
&\mathbb{E}\|\boldsymbol{x}^{(r)}_n-S_n^{(r)}(\boldsymbol{x}^{(r)}_n)\|^2=\sum_{K=0}^{s}\Pr(k_{n}^{(r)}=K)\sum_{i=1}^{s-K}x_i^2
\\& 
\leq\left(1-e^{-\frac {(u+\log_2s)}{A_n c_n}}\right)\sum_{K=0}^{s-1}e^{-\frac {K(u+\log_2s)}{A_n c_n}}\frac{s-K}s\|\boldsymbol{x}^{(r)}_n\|^2
\\&\leq\left(1- e^{-\frac {(u+\log_2s)}{A_n c_n}} \right)\|\boldsymbol{x}^{(r)}_n\|^2.\notag
\end{aligned}
\end{equation}

This completes the proof. \hfill$\square$

\section{Proof of Lemma 3}
\label{lemma3}
Let $r'$ denote the nearest round in which device $n$ establishes a connection with the MES after round $r$. There is $\theta^{(r)}_n \leq \theta^{(r')}_n$ and $\mathbb{E}\left[\left(\theta^{(r)}_n\right)^2\right] \leq \mathbb{E}\left[\left(\theta^{(r')}_n\right)^2\right]$. 

The second moment of model staleness of $n$ in round $r'$ is
\begin{equation}
\begin{aligned}
&\mathbb{E}\left[\left(\theta^{(r')}_n\right)^2\right]=\sum_{\theta=1}^\infty\theta^2\Pr\left(\theta^{(r')}_n=\theta\right).\label{etheta1}
\end{aligned}
\end{equation}
If device $n$ contacts the MES in two consecutive rounds, $\theta^{(r')}_n=1$. If device $n$ contacts the MES in round $r'$ and $\theta^{(r')}_n>1$, device $n$ and the MES are in the inter-contact period at the beginning of the next round $r'+1$. We use the symbol $\chi$ to denote this event and $t_n^{(r')}$ to denote the remaining duration of the inter-contact time from the beginning of the round $r'+1$. Therefore, we have 
\begin{equation}
\begin{aligned}
\Pr\left(\theta^{(r')}_n=\theta\right) = \Pr\left(\chi \right) \Pr\left(\theta\delta\leq t_n^{(r')} \leq(\theta+1)\delta | \chi \right) \label{preq}
\end{aligned}
\end{equation}
Due to the memoryless property of the exponential distribution, $t_n^{(r')}$ also follows an exponential distribution with mean $\lambda$. Also, based on the renewal theorem, we have $\Pr(\chi)=\frac{\lambda}{\lambda+c} $.
Plugging (\ref{preq}) into (\ref{etheta1}), we have
\begin{equation}
\begin{aligned}
\mathbb{E}\left[\left(\theta^{(r')}_n\right)^2\right] &\leq1+\sum_{\theta=2}^\infty\theta^2\Pr\left(\chi \right) \Pr\left(\theta\delta\leq t_n^{(r')} \leq(\theta+1)\delta\right) \\
&=1+\frac{\lambda}{\lambda+c}\sum_{\theta=2}^\infty\theta^2\int_{\theta\delta}^{(\theta+1)\delta}\frac{e^{-\frac{t_n^{(r')}}\lambda}}\lambda d t_n^{(r')} \\
&=1+\frac{\lambda}{\lambda+c} \cdot \frac{e^{-\frac{4\delta}{\lambda}}-3e^{{-\frac{3\delta}{\lambda}}}+4e^{{-\frac{2\delta}{\lambda}}}}{1-2e^{{-\frac{\delta}{\lambda}}}+e^{{-\frac{2\delta}{\lambda}}}}.\notag
\end{aligned}
\end{equation}

\section{Proof of Theorem 2}
\label{theorem2}

To prove Corollary 1, we first derive the following lemma to bound the local memory.

\noindent\textbf{Lemma 4.} \emph{The second moment of the local memory of every device $n\in\mathcal{N}$ in every round $r \in \mathcal{R}$ is bounded by}
\begin{equation}
\begin{aligned}
\mathbb{E}\left\|\boldsymbol{e}_n^{(r)}\right\|^2 \leq \frac{4(1-\gamma_n^2)}{\gamma_n^2}\Theta_n\eta^2 G^2,
\end{aligned}
\end{equation}
\emph{where the expectation is taken over the randomness of contact time, inter-contact time and SGD.}

\noindent \emph{Proof:} This is an extension of Lemma 2 in [46]. The proof is provided
in [58]. \hfill$\square$ 

Now we can turn to prove Theorem 2. We rewrite (\ref{theorem11}) as 
\begin{equation}
\begin{aligned}
&\mathbb{E}\left\|\nabla F(\boldsymbol{z}^{(R)})\right\|^2 \leq \frac{4\left(F(\boldsymbol{w}^{(0)})  - F(\boldsymbol{w}^{*})\right)}{R}
\\&+ \frac{4 L^2}{NR}\sum_{r=0}^{R-1}\sum_{n\in \mathcal{N}}\left\|\boldsymbol{e}_n^{(\kappa_n^{(r)})}\right\|^2+\left\|S(\boldsymbol{x}_n^{(\kappa_n^{(r)})})\right\|^2\\
&+\frac{8 \eta^2 L^2 G^2}{NR}\sum_{r=0}^{R-1}\sum_{n\in \mathcal{N}}\left(\theta_n^{(r)}\right)^2+\frac{4\eta L\sigma}{N}.\notag
\end{aligned}
\end{equation}
Based on Lemmas 2, 3, and 4, taking an expectation over contact and inter-contact times, and setting $\eta = \frac{1}{2L\sqrt{R}}$, we prove Theorem 2.
\begin{equation}
\begin{aligned}
&\mathbb{E}\left\|\nabla F(\boldsymbol{z}^{(R)})\right\|^2 \leq \frac{8\left(F(\boldsymbol{w}^{(0)})  - F(\boldsymbol{w}^{*})\right)}{\sqrt{ R}}+\frac{2 \sigma}{N\sqrt{R}}
\\& +\frac{ G^2}{NR}\sum_{n\in \mathcal{N}} \frac{(16-8
\gamma_n-11\gamma_n^2+6\gamma_n^3)\Theta_n}{\gamma_n^2}.\notag
\end{aligned}
\end{equation}
\section{Proof of Theorem 3}
\label{lyp}
We define a quadratic Lyapunov function as $L^{(r)} \triangleq\frac{1}{2} \sum_{n\in \mathcal{N}}(q_{n}^{(r)})^2$,
and the Lyapunov drift of a single round as
\begin{align}
\label{lypdrift} 
    &\Delta^{(r)}\triangleq L^{(r+1)}-L^{(r)} = \frac{1}{2} \sum_{n\in \mathcal{N}} \left( (q_{n}^{(r+1)})^2 - (q_{n}^{(r)})^2 \right)\\ 
    &\leq\frac{1}{2} \left( \sum_{n\in \mathcal{N}}\left(q_{n}^{(r)}+\epsilon_{n}^{(r)}\right)^2-(q_{n}^{(r)})^2 \right) \leq \Phi+\sum_{n\in \mathcal{N}} q_{n}^{(r)} \epsilon_{n}^{(r)} ,\notag
\end{align}
where $\epsilon_n^{(r)} \triangleq  E^{(r)}_n- \frac{E^{\text{con}}_n}{R}$, $\phi_n \triangleq \max_r\{|\epsilon_n^{(r)}|\}$, and $\Phi \triangleq \sum_{n\in \mathcal{N}}(\phi_n)^2$.
By adding $-\sum_{n \in \mathcal{N}} V
U_{n}^{(r)} $ on both sides of (\ref{lypdrift}), an upper bound on the single-round drift-plus-penalty function is given by
\begin{equation}
\begin{aligned}
&\Delta(t) -\sum_{n \in \mathcal{N}} V
U_{n}^{(r)} \leq \Phi+ \sum_{n\in \mathcal{N}} q_{n}^{(r)} \epsilon_{n}^{(r)}   - V
U_{n}^{(r)}.\notag
\end{aligned}
\end{equation}
We define the overall drift as $\Delta \triangleq  \Delta(R)-\Delta(1) = \frac{1}{2} \sum_{n\in \mathcal{N}}(q_{n}^{(R)})^2$. Then, the overall drift-plus-penalty function is bounded by:

\begin{align}
&\Delta - \sum_{r=0}^{R-1}\sum_{n \in \mathcal{N}} V
 U_{n}^{\dagger(r)}
 \label{lyp1} \leq R  \Phi+ \sum_{r=0}^{R-1} \sum_{n\in \mathcal{N}} q_{n}^{(r)} \epsilon_{n}^{\dagger(r)} - V
 U_{n}^{\dagger(r)}\notag\\
& \overset{(a)}{\leq} R  \Phi+ \sum_{r=0}^{R-1} \sum_{n\in \mathcal{N}} q_{n}^{(r)} \epsilon_{n}^{*(r)}  - V
 U_{n}^{*(r)} .
\end{align}
where inequality $(a)$ holds because solving $P1$ yields a minimum value of (\ref{obj1}).
Based on the definition of $q_{n}^{(r)}$, we have $q_{n}^{(r+1)} - q_{n}^{(r)} \leq \phi_n, \forall n \in \mathcal{N},r \in \mathcal{R}$, and therefore
\begin{equation}
\begin{aligned}
q_{n}^{(r)} \epsilon_{n}^{*(r)} = (q_{n}^{(r)}-q_{n}^{(1)}) \phi_n \leq (r-1) (\phi_n)^2.\label{lyp2}
\end{aligned}
\end{equation}
Substituting (\ref{lyp2}) into (\ref{lyp1}), we have
\begin{equation}
\begin{aligned}
&\Delta- \sum_{r=0}^{R-1}\sum_{n \in \mathcal{N}} V
U_{n}^{\dagger(r)} \leq R^2 \Phi - \sum_{r=0}^{R-1}\sum_{n \in \mathcal{N}} V
 U_{n}^{*(r)}.\notag
\end{aligned}
\end{equation}
Since $\Delta\geq 0$, we have
\begin{equation}
\begin{aligned}
\sum_{r=0}^{R-1}\sum_{n \in \mathcal{N}} 
 U_{n}^{\dagger(r)} \geq   \sum_{r=0}^{R-1}\sum_{n \in \mathcal{N}} 
U_{n}^{*(r)}- \frac{R^2 \Phi}{V}.\notag
\end{aligned}
\end{equation}
For energy consumption, we have
\begin{equation}
\begin{aligned}
&\sum_{r=0}^{R-1} \left( E^{(r)}_n - \frac{E^{\text{con}}_n}{R} \right)  \leq \sum_{r=0}^{R-1} q_{n}^{(r+1)} - q_{n}^{(r)} \\& \leq \sqrt{2\Delta} \leq \sqrt{2R^2 \Phi - \sum_{r=0}^{R-1}\sum_{n \in \mathcal{N}} 2V  U_{n}^{*(r)}}.\notag
\end{aligned}
\end{equation}
Theorem 3 is proved.
\section{Proof of Proposition 2}
\label{Proposition2}
Then the KKT condition of $P3$ is given by:
\begin{equation}
\begin{aligned}
& \nu_n^{(1)*} p_n^{*(r)} = 0,\\
& \nu_n^{(2)*} ( p_n^{*(r)} - p^{\text{max}}_n) = 0,\\
& \nu_n^{(3)*} \left(p_n^{*(r)}-\frac{B_n N_0}{|h_n^{(r)}|^2}\left(2^ {\frac{s \left(u + \log_2s \right)}{\tau^{(r)}_n B_n}}-1 \right) \right) = 0,\\
& \nu_n^{(1)*}, \nu_n^{(2)*}, \nu_n^{(3)*}\geq 0,\\
& -\frac{\frac{ 3 V \zeta_n^{(r)} \theta_n^{(r)} 
 \tau^{(r)}_n B_n \left\|\boldsymbol{x}_n^{(r)}\right\|^2}{s \left(u + \log_2s \right)}\frac{|h_n^{(r)}|^2}{B_n N_0}}{1+\frac{p_n^{*(r)}|h_n^{(r)}|^2}{B_n N_0}} + \tau^{(r)}_n  q_{n}^{(r)}\\& -\nu_n^{(1)*} + \nu_n^{(2)*} + \nu_n^{(3)*} = 0. \notag
\end{aligned}
\end{equation}
If two of $\nu_n^{(1)*}$, $\nu_n^{(2)*}$ and $\nu_n^{(3)*}$ are non-zero, there is no solution to these equations. Therefore, four cases are considered:\\
1) If $\nu_n^{(1)*},\nu_n^{(2)*},\nu_n^{(3)*} = 0$, then
$$p_n^{*(r)} = \frac{3 V \zeta_n^{(r)} \theta_n^{(r)} 
  B_n \left\|\boldsymbol{x}_n^{(r)}\right\|^2}{q_{n}^{(r)}s(u + \log_2s)  } - \frac{B_n N_0}{|h_n^{(r)}|^2}.$$\\
2) If $\nu_n^{(1)*} \neq 0$, $\nu_n^{(2)*}, \nu_n^{(3)*} = 0$, then
$p_n^{*(r)} = 0$.\\
3) If $\nu_n^{(2)*} \neq 0$, $\nu_n^{(1)*}, \nu_n^{(3)*} = 0$, then $p_n^{*(r)} = p^{\text{max}}_n$.\\
4) If $\nu_n^{(3)*} \neq 0$, $\nu_n^{(1)*}, \nu_n^{(2)*} = 0$, then $$p_n^{*(r)} = \frac{B_n N_0}{|h_n^{(r)}|^2}\left(2^ {\frac{s \left(u + \log_2s \right)}{\tau^{(r)}_n B_n}}-1 \right).$$ Combining these, we prove Proposition 2.


\end{document}